\documentclass{article}
\usepackage[margin=2cm]{geometry}
\usepackage{graphicx}
\usepackage{booktabs}
\usepackage{tabularx}
\usepackage{geometry}
\usepackage{colortbl}
\usepackage{xcolor}
\usepackage{authblk}

\newcommand{\TN}{{\textsf{TinyStories}}}

\newcommand{\TNI}{{\textsf{TinyStories-Instruct}}}

\newcommand{\red}[1]{\textcolor{red}{#1}} 

\title{\TN: How Small Can Language Models Be and Still Speak Coherent English?}

\author{Ronen Eldan\thanks{roneneldan@microsoft.com}~}
\author{Yuanzhi Li\thanks{yuanzhili@microsoft.com}}
\affil{Microsoft Research}
\date{April 2023}
\usepackage{comment,color,soul}
\begin{document}
\maketitle
\begin{abstract}
Language models\cite{brown2020language,bubeck2023sparks,openai2023gpt4} (LMs) are powerful tools for natural language processing, but they often struggle to produce coherent and fluent text when they are \textbf{small}. Models with around 125M parameters such as GPT-Neo (small)~\cite{gpt-neo} or GPT-2 (small)~\cite{radford2019language} can rarely generate coherent and consistent English text beyond a few words even after extensive training. This raises the question of whether the emergence of the ability to produce coherent English text only occurs at larger scales (with hundreds of millions of parameters or more) and complex architectures (with many layers of global attention). 

In this work, we introduce \textbf{\TN}, a synthetic dataset of short stories that only contain words that a typical 3 to 4-year-olds usually understand, generated by GPT-3.5 and GPT-4. We show that \TN~can be used to train and evaluate LMs that are much smaller than the state-of-the-art models (\textbf{below 10 million total parameters}), or have much simpler architectures (\textbf{with only one transformer block}), yet still produce fluent and consistent stories with several paragraphs that are diverse and have almost perfect grammar, and demonstrate reasoning capabilities.

We also introduce a new paradigm for the evaluation of language models: We suggest a framework which uses GPT-4 to grade the content generated by these models as if those were stories written by students and graded by a (human) teacher. This new paradigm overcomes the flaws of standard benchmarks which often require the model's output to be very structured, and moreover it provides a multidimensional score for the model, providing scores for different capabilities such as grammar, creativity and instruction-following.

We hope that \TN~can facilitate the development, analysis and research of LMs, especially for low-resource or specialized domains, and shed light on the emergence of language capabilities in LMs.
\end{abstract}
\section{Introduction}
Natural language is rich and diverse. It is not only a system of rules and symbols, but also a way of conveying and interpreting meaning~\cite{winograd1972understanding}. To understand and produce language, one needs not only to master the technical rules of grammar and knowledge of vocabulary, but also to have sufficient factual information and to be able to reason logically and contextually. Therefore, autoregressive language models, which are able to generate coherent English text, must have acquired some degree of these capabilities as well. For example, consider the following incomplete sentence:
\begin{quote}
Jack was hungry, so he went looking for $\langle \_ \rangle$
\end{quote}
To complete this sentence in a sensible way, the language model needs to know that hunger is a state that motivates people to seek food, and that food is a category of things that can satisfy hunger. It also needs to choose a word that fits the syntactic and semantic constraints of the sentence (such as ``a snack''), and that is plausible given the situation and the background knowledge. 

An example that illustrates the need for reasoning is:
\begin{quote}
Lily wanted to get either a cat or a dog. Her mother didn't let her get a dog so instead she $\langle \_ \rangle$
\end{quote}
To complete this sentence, the language model needs to invoke \emph{reasoning}: it needs to apply the principle of disjunction elimination: if Lily wants either a cat or a dog, and she cannot get a dog, then she must choose a cat. It also needs to choose a words that expresses Lily's intention or action that is coherent with the tone and style of the text.

Language models have been shown to exhibit a range of emergent abilities, such as summarization, arithmetic, translation, and commonsense reasoning, as they are scaled up in size and trained on diverse and large corpora ~\cite{raffel2020exploring,brown2020language,bubeck2023sparks,openai2023gpt4}. These abilities suggest that language models are not only learning the surface patterns of language, but also acquiring some degree of semantic and logical understanding of the world and the text. However, it is not clear at what scale these abilities emerge, and how they depend on the model architecture and the data distribution.

Perhaps the most fundamental ability for a language model is to produce coherent and fluent English text, which, as we discussed above, requires not only grammatical and lexical knowledge, but also factual information and contextual reasoning. How well can language models generate text that is consistent, diverse, and meaningful? And what are the minimal requirements for a language model to achieve this ability?

So far, the evidence points to the fact that producing coherent text already requires quite a large scale: small language models (SLMs) are very limited in their performance and capabilities, especially in text generation tasks. For example, models with around 125M parameters such as GPT-Neo (small) or GPT-2 (small) can rarely generate any consistent text beyond a few words even after extensive training on large corpora such as the Pile~\cite{gao2020pile}, Common Crawl~\cite{commoncrawl} or the CC-100~\cite{wenzek2019ccnet}. These models often produce incoherent, repetitive, or nonsensical sentences, and fail to maintain a clear topic or a logical structure across paragraphs~\cite{holtzman2019curious}. This raises the question of whether the emergence of the ability to speak coherent English requires large models (with hundreds of millions of parameters or more) and complex architectures (with many layers of global attention).

However, it is currently not clear whether the inability of SLMs to produce coherent text is a result of the intrinsic complexity of natural language, or of the excessive breadth and diversity of the corpora used for training. When we train a model on Wikipedia, for example, we are not only teaching it how to speak English, but also how to encode and retrieve an immense amount of facts and concepts from various domains and disciplines. Could it be that SLMs are overwhelmed by the amount and variety of information they have to process and store, and that this hinders their ability to learn the core mechanisms and principles of language?

\textbf{This raises the question of whether we can design a dataset that preserves the essential elements of natural language, such as grammar, vocabulary, facts, and reasoning, but that is much smaller and more refined in terms of its breadth and diversity.} Such a dataset would allow us to isolate and examine the minimal requirements for a language model to generate coherent and fluent text, and to evaluate its performance and capabilities more precisely and fairly. Moreover, such a dataset would facilitate the development and analysis of SLMs, especially for low-resource or specialized domains, where large and diverse corpora are either unavailable or undesirable.

In this paper, we introduce \TN\footnote{The dataset is available on Huggingface named \TN.}, a synthetic dataset of short stories that are intended to contain only words that most 3 to 4-year-old children would typically understand, generated by GPT-3.5 and GPT-4. \TN~is designed to capture the essence of natural language, while reducing its breadth and diversity. Each story consists of 2-3 paragraphs that follow a simple plot and a consistent theme, while the whole dataset aims to span the vocabulary and the factual knowledge base of a 3-4 year old child \\

Based on this dataset, our paper makes several main contributions:
\begin{itemize}
\item
\textbf{Our main contribution} is that we show \TN~can be used to train and evaluate SLMs\footnote{Our models are available on Huggingface named \TN-1M/3M/9M/28M/33M/1Layer/2Layer and \TNI-$*$. We use GPT-Neo architecture with window size 256 and context length 512. We use GPT-Neo tokenizer but only keep the top 10K most common tokens.} that are much smaller than the state-of-the-art models (below 10 million parameters with an embedding dimension of 256), or have much simpler architectures (with only one transformer block), yet still produce \textbf{a diverse set of fluent and consistent stories} that are comparable or superior to those generated by larger and more complex models. Moreover, despite of the small size of the models, we still observe an \textbf{emergence of reasoning capabilities, knowledge of general facts and ability to follow certain instructions}. 
\item 
We introduce a new paradigm for evaluating language models using GPT-4, which overcomes many of the limitations of standard benchmarks.
\item
We show that although the training of generative models on \TN~can typically be done in less than a day on a single GPU, they still exhibit many behaviors similar to the ones observed in LLMs, such as scaling laws, trade-offs between width and depth, etc. Even with limited computational resources, we are able to conduct extensive experiments to study the effects of different hyperparameters, architectures and training methods on the performance and quality of the models.
\item
We show that the trained SLMs appear to be substantially more \textbf{interpretable} than larger ones. When models have a small number of neurons and/or a small number of layers, we observe that both attention heads and MLP neurons have a \textbf{meaningful} function: Attention heads produce very clear attention patterns, with a clear separation between local and semantic heads, and MLP neurons typically activated on tokens that have a clear common role in the sentence. We visualize and analyze the attention and activation maps of the models, and show how they relate to the generation process and the story content.
\end{itemize}

To give the reader a first impression of the abilities of models trained on \TN, we compare the completion of a 28M parameter model trained on \TN\footnote{For the sake of replicability, most completions which appear in this paper, including this one, were generated with zero temperature.}~with that of GPT2-XL, which is two orders of magnitude bigger (1.5B parameters), on a sample prompt\footnote{This prompt was composed manually and then verified to have no 6-gram overlap with the dataset.} in Figure \ref{fig:comp1}. We remark that the architectures and training scheme of the models are essentially the same.

\begin{figure}[h!]
\centering
\scriptsize
\begin{tabular}{p{0.18\textwidth}p{0.75\textwidth}}
\toprule
\textbf{Prompt} & Tom and Jane are friends. One day, Jane goes to Tom's house. Tom has a big pot of soup. He wants to share it with Jane. ``Jane, do you want some soup?'' Tom asks. ``Yes, please. It looks yummy,'' Jane says. Tom pours some soup into two bowls. He gives one bowl to Jane. Jane takes a spoonful of soup, but then she makes a face. The soup is \\
\midrule
Completion by GPT2-XL (\textbf{1.5B} parameters) & 
too old. Tom scowls.

``Don't be that way! Your soup has to be fresh like your daughter,'' Tom says, angry and disappointed. ``Bring me some more,'' Jane cries. ``No! Go away, I like my soup old and fresh!''\\
\midrule
Completion by a \textbf{28M}-parameters model trained on \TN&
very bitter. She does not like it. She says, ``I don't like this soup. It is too bitter.''

Tom is sorry. He says, ``I'm sorry, Jane. I didn't know you don't like bitter soup. I will make you something else.'' He looks around the kitchen and finds some bread and cheese. He puts them on the table and says, ``Here, Jane. You can have some bread and cheese. They are not bitter. They are sweet and yummy.''

Jane is happy. She says, ``Thank you, Tom. You are a good friend. I like bread and cheese. They are not bitter.'' \\
\bottomrule
\end{tabular}
\caption{Comparison between the completions given by a 28M-parameters model trained on the \TN~dataset and GPT-2XL (1.5B)} \label{fig:comp1}
\end{figure}

Returning to the examples given at the beginning of the introduction, we highlight the completions in Figure \ref{fig:frc1}. Those completions, along with many other examples given throughout the paper, demonstrate that even very small models (2.5M) or models with only one transformer layer are able to attain factual knowledge, and that slightly bigger models are able to express a certain extent of reasoning. The comparison with the performance of GPT2-XL reveals that the \TN~dataset allows us to exhibit the emergence of these capabilities at significantly smaller scales compared to models trained on large language corpora. More evidence for this will be provided throughout the rest of the paper.

\begin{figure}[h!]
\centering
\scriptsize
\begin{tabular}{p{3cm}p{2cm}p{2cm}p{2cm}p{2cm}}
\toprule
\textbf{Prompt} & \textbf{2.5M, 8 layers} & \textbf{33M, 4 layers} & \textbf{21M, 1 layer} & \textbf{GPT2-XL (1.5B)} \\
\midrule
Alice was so tired when she got back home so she went & to bed. &  straight to bed. & to bed with a big smile on her face. & outside\\
\midrule
Lily likes cats and dogs. She asked her mom for a dog and her mom said no, so instead she asked &   her mom if she could have a dog. &   her dad for a cat.  &   her mom again.  &  her dad. They brought her a Chihuahua, and she took to them immediately. \\ 
\midrule
Alice and Jack walked up the street and met a girl in a red dress. The girl said to them, "Hi, I'm Jane. What are your names?" & Alice smiled and said, "My name is Daisy. What's your name?" &  Alice said, "I'm Alice and this is Jack." & Jack smiled and said, "I'm Jane. Nice to meet you!" &  Jane said, "Jack and Kate"\\
\bottomrule
\end{tabular}
\caption{Example performance of different models on factual knowledge, reasoning and contextual prompts. The first three models were trained on the \TN~dataset.} \label{fig:frc1}
\end{figure}

\section{Description of the \TN~dataset}

As mentioned above, the idea behind the \TN~dataset is to have a corpus that combines all the qualitative elements found in natural language, such as grammar, vocabulary, facts, and reasoning, but is also smaller, less diverse, and more restricted in terms of its content. A natural approach to achieve this is to use the fact that young children qualitatively attain the aforementioned intellectual abilities, albeit with much less exposure to language than an adult human~\cite{flavell1990young,miller1964development}.

To this end, we rely on the latest text generation models by OpenAI (GPT-3.5 and GPT-4) which are able to produce large amounts of synthetic content according to instructions. In particular, we instruct the models to produce content that only uses vocabulary that a typical 3-year-old child would understand. We restrict the content to have the format of short stories in English. The main challenge in using large language models for producing training data is generating a dataset that is sufficiently \textbf{diverse}: prompting those models to produce stories, even if the temperature of generation is set to a high value, will still produce a very repetitive dataset, whose diversity is very far from what is required for training a language model that has a comparable "understanding" of language to that of children.

In order to address the problem of creating a diverse dataset, we collected a vocabulary consisting of about 1500 basic words, which try to mimic the vocabulary of a typical 3-4 year-old child, separated into nouns, verbs, and adjectives. In each generation, 3 words are chosen randomly (one verb, one noun, and one adjective). The model is instructed to generate a story that somehow combines these random words into the story. As we argue below, this greatly increases the diversity of the dataset, forcing the stories to span the entire vocabulary a child is familiar with, and to include a rich set of ways to combine different concepts. In addition, we constructed a list of possible features a story could have (such as a dialogue, a plot twist, a bad ending or a moral value). For each story we generated a random subset of those features and prompted the model with the extra requirement for the story to have these features.

Here is a sample generation, including the prompt and the story generated by GPT-3.5.

{\scriptsize
\begin{quote}
\textbf{Write a short story (3-5 paragraphs) which only uses very simple words that a 3 year old child would likely understand. The story should use the verb "decorate", the noun "thunder" and the adjective "ancient". The story should have the following features: the story should contain at least one dialogue, the story has a bad ending. Remember to only use simple words! }\\

{\it Once upon a time, in an ancient house, there lived a girl named Lily. She loved to decorate her room with pretty things. One day, she found a big box in the attic. She opened it and saw many shiny decorations. Lily was very happy and decided to use them in her room. \\
As Lily was decorating her room, the sky outside became dark. There was a loud thunder sound, and Lily got scared. She ran to her mom and said, "Mommy, the thunder is so loud!" Her mom hugged her and said, "Don't worry, it will pass soon." \\
But the thunder did not stop. It got louder and louder, and the ancient house started to shake. Suddenly, the roof fell down on the pretty decorations. Lily was sad because her room was not pretty anymore. The end.}
\end{quote}}

\subsection{\TNI: An instruction-following variant of the dataset}
Another challenge of natural language generation is to produce texts that are coherent, relevant, and consistent with given instructions or constraints. To evaluate the capability of generative models in terms of instruction-following rather than just content creation, we created a variant of \TN\footnote{Available on Huggingface under the name TinyStoriesInstruct.}~(referred to as \TNI) in which each story is preceded by a corresponding set of instructions. The instructions are of four types:

\begin{enumerate}
    \item A list of words to be included in the story.
    \item A sentence that should appear somewhere in the story.
    \item A list of features (possible features: dialogue, bad ending, moral value, plot twist, foreshadowing, conflict).
    \item A short summary (1-2 lines) of the story.
\end{enumerate}

Every entry in this dataset has a random subset of the these instructions followed by the story itself. This way, given a model trained on this dataset, we can compose a prompt to reflect any set of instructions and use the model to generate a corresponding story. Here is an example from the dataset:
{\scriptsize
\begin{quote}
\textbf{Summary}: Lily and Timmy build a sandcastle together and learn to compromise, but it gets knocked over by a gust of wind. They find beauty in the broken sandcastle and play happily with a butterfly.\\
\textbf{Features}: Dialogue, Foreshadowing, Twist\\
\textbf{Sentence}: One day, she went to the park and saw a beautiful butterfly.\\
\textbf{Words}: disagree, network, beautiful\\
\textbf{Story}: Once upon a time, there was a little girl named Lily. She loved to play with her toys and her friends. One day, she went to the park and saw a beautiful butterfly. She tried to catch it, but it flew away.
Lily's friend, Timmy, came to play with her. They decided to build a sandcastle, but they disagreed on how to build it. Lily wanted to make it tall, while Timmy wanted to make it wide. They argued for a while, but finally agreed to make it both tall and wide.
As they were building the sandcastle, they noticed a group of ants working together. They realized that the ants were building a network of tunnels. Suddenly, a gust of wind came and knocked over their sandcastle. They were sad, but then they saw the butterfly again. It landed on the remains of their sandcastle and they realized that it was still beautiful, even in its broken state. They smiled and played together happily.
\end{quote}
}
\TNI~was created in the following way: For each story in the original dataset, we already had a list of words and features that were used to create it. In addition, we used GPT-3.5 to create short summaries of our stories, and from each story we extracted a random sentence (which could be any sentence in the story except for the first one). Then we chose a random subset of these and combined them into an instruction, followed by the story itself.

\section{GPT-Eval: A framework for evaluating models trained on \TN}

While the common paradigms for evaluation of language models usually rely on structured evaluation datasets in the form of a task where the output of the model has to match a given answer, we introduce a new paradigm that is arguably more suitable in this context. Again, we take advantage of existing large language models to evaluate the stories generated by the model. We provide the model with a story's beginning, taken from a \textbf{manually-prepared} dataset consisting of around 50 prompts, generate a completion using the model, and provide the story's beginning together with the model's completion to GPT-4, asking it to grade the completion assignment in terms of grammar, creativity, and its consistency with the beginning of the story. 
The prompts usually contain a partial sentence that needs to be completed, where we attempted to write them in a way that it would be a grammatically nontrivial task to complete the sentence that was cut short.

As an example, we took the story that appears in the example above and cut it after the word "loud". We generated a completion of the above story using a model with roughly 28M parameters (the generation is highlighted in yellow), then we prompted GPT-4 to first provide a verbal assessment and then grade the completion. The transcript of interaction is as follows, with the parts produced by GPT-4 in italic font:
{\scriptsize
\begin{quote}
\textbf{the following exercise, the student is given a beginning of a story. The student needs to complete it into a full story. The exercise tests the student\'s language abilities and creativity. The symbol *** marks the separator between the prescribed beginning and the student's completion:}

Once upon a time, in an ancient house, there lived a girl named Lily. She loved to decorate her room with pretty things. One day, she found a big box in the attic. She opened it and saw many shiny decorations. Lily was very happy and decided to use them in her room. \\
As Lily was decorating her room, the sky outside became dark. There was a loud*** \hl{thunderclap and a bright flash of lightning. Lily was a little scared, but she knew she had to be brave. She told her mom and dad, and they all went outside to see what was going on.

When they got outside, they saw a big storm coming. The wind was blowing hard, and the rain was pouring down. Lily, her mom, and her dad knew they had to find a safe place to hide. They found a big tree and hid under it. The storm passed, and the sun came out again. Lily, her mom, and her dad were all safe and warm inside their ancient house.} \\ \\
\textbf{Please provide your general assessment about the part written by the student (the one after the *** symbol). Is it gramatically correct? Is it consistent with the beginning of the story? Pay special attention to whether the student manages to complete the sentence which is split in the middle by the separator ***.}

{\it The student's completion of the story is mostly consistent with the beginning of the story. It maintains the focus on Lily and her family, and the sentence split by the separator is completed correctly. However, the student's addition does not fully integrate the shiny decorations found in the attic, which were a significant part of the beginning.

The grammar is generally correct, but there are a few minor errors: $\langle \textrm{list omitted} \rangle$.

Overall, the student's completion of the story demonstrates adequate language abilities and creativity, but could benefit from better integration of the shiny decorations and minor grammar improvements.}

\textbf{Now, grade the student's completion in terms of grammar, creativity, consistency with the story's beginning and whether the plot makes sense. Moreover, please provide your best guess of what the age of the student might be, as reflected from the completion. Choose from possible age groups: A: 3 or under. B: 4-5. C: 6-7. D: 8-9. E: 10-12. F: 13-16.} \\

{\it Grammar: 8/10, Creativity: 7/10, Consistency: 7/10, Age group: E (10-12) }
\end{quote}
}

\begin{figure}[h!] 
\centering
\includegraphics[width=0.78\textwidth]{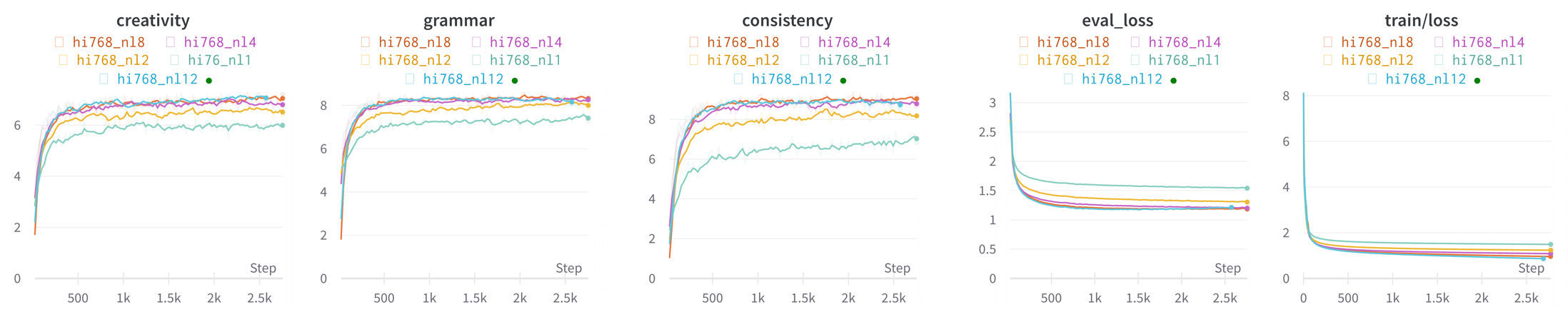}
\caption{Evaluation loss and the GPT-Eval scores during training for the GPT-neo models with embedding dimension 768 and different number of layers. We can see that the GPT-4 evaluation scores increase as evaluation losses decrease.} \label{fig:trainingscores}
\end{figure}

To perform the full evaluation, for each of the manually-constructed prompts in the evaluation set, we use the trained model to generate 10 completions with temperature 1. We average the GPT-4 evaluation scores of all the completions. Figure \ref{fig:trainingscores} provides an example of the evolution of the different scores as the training progresses, along with the training and evaluation loss, for several model sizes. Figure \ref{tab:merged} shows how the different scores vary with different model sizes and architectures, after a fixed number of training steps.

Our evaluation method for models trained on \TNI~also relies on GPT-4. We have implemented a corresponding evaluation pipeline, with an evaluation dataset containing 50 different sets of instructions to create a story (we verified that these are disjoint from the training set). In the scoring phase, we provide GPT-4 with both the instructions and with the generated story. We prompt GPT-4 to base the consistency score on the extent to which the story accurately reflects the given instruction. In addition, we added a \textbf{Plot} category that reflects the extent to which the plot is coherent. Figure \ref{fig:evalinstr} illustrates the whole pipeline that combines the generation of the story by our model, and its evaluation by GPT-4. Scores assigned to models of different sizes appear in the two right-hand columns of the table in Figure \ref{tab:merged}.

\subsection{First insights that arise from our evaluation method}
\begin{figure}[h!]
\includegraphics[width=1\textwidth]{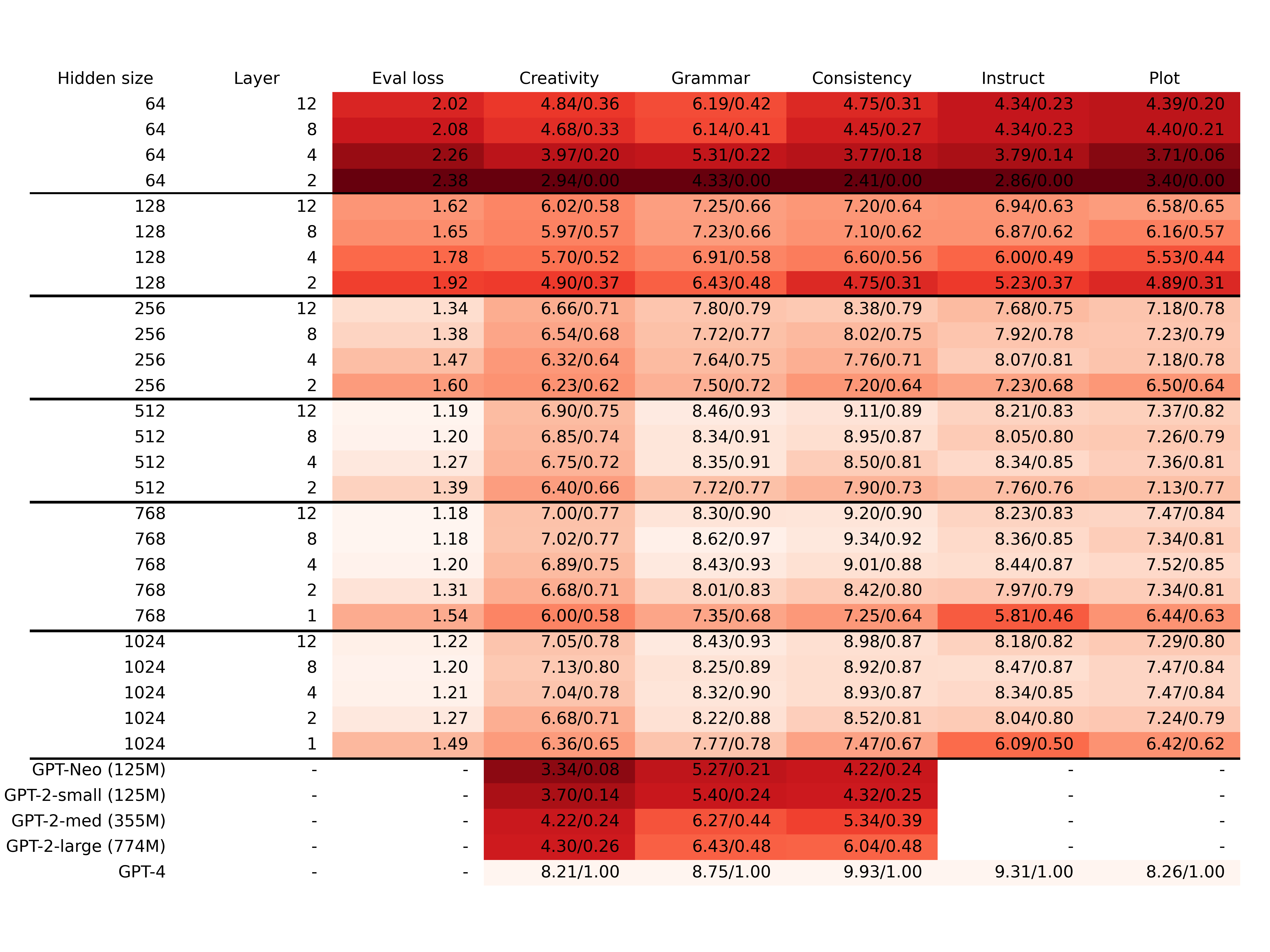}
\caption{Evaluation results of different hidden sizes and layers for story generation and Consistency (here we use format $a/b$, $a$ means the original score, $b$ means the normalized score according to $(a - a_{\min})/(a_{\max}-a_{\min})$). For GPT-Neo and GPT-2 generations, we truncate the result when there's a repeating 4-gram (after the point the models will just repeat the same sentences over and over). On the other hand, our models learn when to stop generating correctly.} \label{tab:merged}
\end{figure}

Our proposed evaluation method gives a way to obtain a more fine-grained assessment of the model, due to which we can draw conclusions regarding the dependence of different types of capabilities on the size and architecture of the model. While all the evaluation scores are consistently increasing with the decrease of evaluation loss, a more careful scrutiny of the results reveals the following:
\begin{itemize}
\item
Figure~\ref{fig:trainingscores} suggests that shallower models perform better in terms of grammar compared to content consistency, meaning that model depth is more important for keeping consistent with the content than for generating syntactically correct language (we provide additional evidence for this in the next section).
\item
In the same figure, we observe that the score for grammar plateaus at an earlier stage than the other two scores. Furthermore, in Table~\ref{tab:merged}, we also see that while grammar can be mastered by relatively small models, consistency and creativity only emerge at a larger size. 
\item
Table~\ref{tab:merged} further suggests that the ability to generate a completion that is consistent with the beginning of the story \textbf{emerges} when the hidden size of the model increases from 64 to 128. 
\item
We also see that the largest model that we have trained on \TN~(with roughly 80M parameters) reaches almost perfect scores in terms of grammar and consistency. However, it falls short of GPT-4's abilities in terms of creativity quite significantly, suggesting that creativity continues to improve more substantially with the sizes of the model and dataset, compared to grammar and consistency.
\item
The right-hand columns of Table~\ref{tab:merged} suggests that the models that have only 1 layer seem to struggle quite substantially with following instructions (which likely heavily relies on global attention), and 2 layers seem to be sufficient for a certain extent of instruction-following. Comparing the "Instruct" and "Plot" scores we also see that the quality of instruction-following depends more heavily on the number of layers, in comparison with the coherence of the plot for which the hidden dimension is more important.
\end{itemize}

\begin{figure}[h!]\label{evaltask}
\noindent\fbox{ \parbox{\textwidth}{
{\scriptsize
\textbf{Summary}: A cat performs a new trick for her friends but starts shivering. Her friends give her a warm hug and she realizes their love is the best trick of all. \\
\textbf{Sentence}: The cat started to do her trick, but then something unexpected happened. \\
\textbf{Words}: perform, cold, beautiful. \\
\textbf{Story}: \hl{Once upon a time, there was a beautiful cat. The cat liked to perform tricks for her friends. She would jump, spin, and dance. Her friends loved to watch her perform. \\
One day, the cat went outside to play. It was very cold. The cat did not like the cold, but she still wanted to perform for her friends. She put on her warm coat and hat. The cat started to do her trick, but then something unexpected happened. The cat began to shiver! Her friends saw this and quickly gave her a warm hug. The cat realized that her friends loved her even when she was cold. From that day on, the cat performed tricks for her friends, and they all lived happily ever after.} \\ \\
\textbf{Your task is to evaluate the performance of a student. The student is given the following exercise: \\
Write a short story. The story should contain the words: perform, cold, beautiful. The story should contain the following sentence: [\dots]. The general plot of the story should be: A cat performs [\dots] (feel free to expand beyond this general plot). \\
The student wrote the following story: [\dots] \\ \\
Please provide your general assessment about the story written by the student. Is it gramatically correct? Is it consistent with the requirements in the exercise? } \\
{\it The story written by the student is gramatically correct and consistent with the requirements in the exercise. The student used the words perform, cold, and beautiful, and the sentence and the general plot that were given. The story is simple but coherent and has a positive message. The student could improve the story by adding some details or descriptions to make it more interesting and engaging, such as: [...]} \\
\textbf{Now, grade the story in terms of grammar, creativity, consistency with the instructions and whether the plot makes sense. Moreover, please provide your best guess of what the age of the student might be, as reflected from the completion. Choose from possible age groups:[...]. Use the following grade format: [...]} \\ \\
{\it Grammar: 10/10, Creativity: 6/10, Consistency: 10/10, Plot: 8/10, Age group: C. }} }
}
\caption{Generation of a story according to instructions, and its evaluation using GPT-4. The highlighted yellow part was generated by our model (28M parameters) and the italic parts are GPT4-generated.} \label{fig:evalinstr}
\end{figure}

\section{The performance of small models trained on \TN}
In this section, we give some initial examples that illustrate how \TN~gives rise to models of very small size that can generate coherent language and exhibit common-sense knowledge as well as certain extent of reasoning capabilities. We also provide evidence that the generated content is truly diverse, refuting the possibility that the models simply output content that has been "memorized".

Throughout the section, we work with several architectures of models whose size ranges between roughly 1M and 35M parameters, and whose number of layers range between 1 and 8 layers.  All of the models can be trained on a single V100 GPU within at most 30 hours.

\subsection{First examples}
To get a glimpse into the capabilities of (very) small models trained on \TN, let us first inspect a few examples. We start with the following story, taken from the training set.\footnote{Below, there will be many other examples of completions of stories from \textbf{outside} of the training set.}
{\scriptsize
\begin{quote}
{
Sara and Ben are playing in the snow. They make a big snowman with a hat and a scarf. They are happy and laugh.

But then a big dog comes. The dog is angry and barks. He runs to the snowman and bites his hat. Sara and Ben are scared and cry. "Go away, dog! Leave our snowman alone!" Sara shouts. But the dog does not listen. He bites the scarf and the snowman's nose. He shakes his head and makes the snowman fall.

Sara and \textcolor{red}{{\textbf{ [}}} Ben run to their house. They slam the door and lock it. They are safe and hug each other. "Mom, mom, a big dog broke our snowman!" Ben says. Mom comes and hugs them. She is sorry and kisses them. "Don't worry, my loves. We can make another snowman tomorrow. But now you need to get warm and have some hot chocolate. You are cold and wet." She takes them to the kitchen and makes them hot chocolate. She gives them cookies and marshmallows. She tells them to drink and eat in order. Sara drinks first, then Ben. Ben eats first, then Sara. They feel better and smile. They thank mom and tell her they love her. They escape from the big dog and the cold. They are happy and warm.\textcolor{red}{{\textbf{ ]}}}}
\end{quote}
}
We now cut the story roughly in the middle, removing the part inside the square brackets and generate an alternative ending with several different models.

To set a baseline, we begin with the completion generated by \textbf{GPT2-XL}, whose size is roughly 1.5 Billion parameters:
{\scriptsize
\begin{quote}
{\it
\dots Sara and \hl{Ben cry and look at the snowman in the snow. Ben asks to do something. Sara tells him to go away. Her hand tightens on the little red button. The snowman is covered in fur. It is red. It is real hair. It is a real snowman. It looks just like the ones they used to make. Ben has a big smile on his face. He looks at the face he has made. And he walks over and kisses the eyes. He is happy to be alive. He has done what he was going to do. He had saved the world.} }
\end{quote}
}

Now, let us provide the generation given by a model trained on \TN, whose size is about 2.5 million parameters\footnote{This example was generated with temperature 0.5.}:
{\scriptsize
\begin{quote}
{\it \dots Sara and \hl{Ben are very sad and angry. They cry and shout at the dog. "Bad dog! Bad dog! Go away!" Ben says. "Go away, bad dog! Leave us alone!" Sara says. The dog does not go away. He wags his tail and licks their faces. Sara and Ben feel sorry for the dog. They want to make him happy. "Maybe we can make a new snowman for him," Sara says. "Maybe," Ben says. They hug the dog and give him a name. The dog wags his tail and licks their faces. "Maybe he likes us," Sara says. "Maybe he likes us," Ben says. The dog wags his tail and licks their faces. He is happy and friendly.}}
\end{quote}
}
First of all, we remark that the above generation is very different from the original ending of the story, which provides a first evidence that it hasn't simply been memorized (we discuss this much more thoroughly later on). We also see that the generated content is consistent with the beginning in terms of plot, characters and events. Arguably, this continuation is of higher quality than the one provided by GPT2-XL \textbf{despite the size of the model being almost 1000 times smaller}. \\

However this raises the question whether the small models are effectively memorizing a small number of templates and the generation is very limited in terms of diversity, which would be a valid concern at this point. We systematically address this later on, in Section \ref{sec:diversity}. By testing the overlap of the generated stories with the training set and by considering out of distribution generation, we are led to the conclusion that those models do have substantial diversity, and do not rely on memorization.

Next, we consider the completion by a model whose size is roughly 21M parameters, but \textbf{only has one transformer layer} (and consequently only one layer of global attention):
{\fontsize{6}{6}\selectfont
\begin{quote}
{\it \dots Sara and \hl{Ben are very sad. They cry and hug each other. They wish the dog would go away. Then they hear a voice. It is their mom. She comes out of the house and sees them. "What happened here?" she asks. "Why are you crying?" 

Sara and Ben tell her what happened. They tell her about the dog and the snowman. Mom hugs them and says, "Don't worry, my loves. The dog didn't mean to hurt you. He was just curious and playful. He didn't mean to hurt you. He just wanted to play."

Sara and Ben feel better. They say sorry to the dog and the snowman. "Can we still play with the dog?" Sara asks. Mom smiles and nods. She says, "Yes, you can. But you have to be gentle and careful. And you have to share the snowman with us." Sara and Ben nod and say, "OK, Mom. We will." They go inside and have some hot chocolate. They are warm and yummy. They are happy.}}
\end{quote}}
While this completion is far from perfect, it is (arguably) consistent with the beginning in the sense that the model successfully captured the tone and the main elements of the plot.

We acknowledge that the two examples above are not the worst-case scenarios for the models, but they are not extremely rare either. Models of this size can sometimes produce less coherent or plausible completions, but they can also often match or surpass the quality of the ones shown here. However, if we increase the number of parameters by an order of magnitude, we observe that the models \textbf{consistently} generate coherent and relevant continuations. For the sake of replicability, examples from this point on will generated at zero temperature, and \textbf{in addition, the models parameters are provided as supplemental material}. For the sake of replicability, examples below are all generated at zero temperature, and \textbf{in addition the models parameters are provided as supplemental material}.

In order to give the reader an impression of the dependence of the quality of completions on the size of the model, Figures \ref{tab:story1}, \ref{tab:story2} and \ref{tab:story3} each provide different completions for one prompt given by models of different sizes and depths. Each table represents a different prompt, which we have manually composed\footnote{We manually verified that the dataset does not contain any entries which are similar or close to these prompts.}. 

We see that the quality of generation clearly improves as a factor of size, and appears to be consistent with the grades given by the GPT-4 evaluation. The smaller model (64\_8) can barely produce a completion which looks coherent with the beginning of the story, and often repeats itself or makes no sense. As the size increases, the models become more and more coherent, and the grammar becomes better. The models can also generate more diverse and creative endings, and use more details and emotions.

We can also notice that models with a small number of layers have a hard time staying in context, even if they do manage to produce syntactically correct English. This suggests that the model lacks the ability to capture the long-term dependencies and the structure of the story. On the other hand, models with more layers can better maintain the consistency and the logic of the story.

An interesting observation is that in Figure \ref{tab:story2}, even though the completions are generated by different models, they begin in a very similar way (all completions have to do with a little girl coming by and talking to the pumpkin). We point out that the reason for this seems to be that the completions are generated with temperature 0. Roughly speaking, this gives rise to the "most likely" completion. In order to demonstrate that the model is capable of generating a more diverse set of endings to the story, we added a completion with a non-zero temperature. It appears, however, that the quality of completion slightly decays when increasing the temperature\footnote{We do not present more evidence for this claim as it goes beyond the scope of the paper.}.

\subsection{Knowledge, reasoning and context-tracking}
Next, we assess the capabilities of the different models on three additional types of prompts: 
\begin{itemize}
\item
\textbf{Factual prompts}, which test the models' knowledge of common sense facts. 
\item 
\textbf{Reasoning prompts}, which test basic reasoning abilities, such as cause and effect and elimination. 
\item 
\textbf{Consistency (context-tracking) prompts} test the models' ability to maintain coherence and continuity with the given context, such as the names and actions of the characters, the setting and the plot. 
\end{itemize}
We report the generated continuations for each model and prompt in three tables (Figure \ref{fig:factual}, Figure \ref{fig:reasoning} and Figure \ref{fig:consistency}), and color-code them according to their success (green), failure (red), or partial success (yellow).

The results show that as the embedding dimension and the number of layers increase, the performance in regards to all three categories improve. The models with higher embedding dimensions and more layers tend to generate more accurate, relevant, and natural continuations, while the models with lower embedding dimensions and fewer layers tend to generate more nonsensical, contradictory, or irrelevant continuations. For example, the model with 1M parameters and 8 layers fails to answer any factual prompt correctly, and often generates sentences that do not make sense or do not follow the grammar. The model with 33M parameters and 8 layers, on the other hand, answers most prompts, from all three categories, correctly. Comparing to the completions given by GPT2-XL (right hand column), we see that despite its much larger size, its performance in all three categories is worse than some of our models.

One interesting finding is that knowledge of facts seems to rely more on the embedding dimension, whereas for context-tracking the number of layers is more important. For example, the model that has only 1 layer does not get any consistency prompt right, but does get some facts right, whereas the model with embedding dimension 64 does not get any fact right, but manages to maintain consistency several times. This suggests that the embedding dimension is more crucial for capturing the meaning and the relations of words, while the number of layers is more crucial for capturing long-range dependencies in the generation.

\newcolumntype{b}{X}
\newcolumntype{s}{>{\hsize=.3\hsize}X}

\begin{figure}[h!] 
\centering
\scriptsize
\begin{tabularx}{\textwidth}{p{1 cm}p{13 cm}p{2.5cm}}
\toprule
\textbf{Model} & \textbf{Generated text} & \textbf{Scores} \\
\midrule
Prompt & Once upon a time there was a little girl named Lucy. She was very adventurous. She loved to explore the world around her, especially when it was bright and sunny outside. \newline
One day, while exploring the nearby park, Lucy came across a ladder leaning on a wall. She was curious to see what's on top, so she climbed the ladder, but when she reached the top, the ladder fell and she was stuck.\newline
A nearby park ranger noticed her and shouted out, " & \\
\midrule
1M \newline 8 layers & What are you doing here?" \newline
Lucy smiled and said, "I'm going to help!"\newline
The ranger smiled and said, "That's a great idea! Let's go see if we can find out."Lucy was so excited and said, "Yes, let's go!"\newline
So, Lucy and the ranger went to the park. They found lots of fun things and had lots of fun. & Grammar: 6/10\newline
Creativity: 3/10\newline
Consistency: 2/10 \\
\midrule
2.5M \newline 8 layers & What are you doing, Lucy?"\newline
Lucy replied, "I'm trying to get out of the ladder, but I can't get out!"\newline
The ranger smiled and said, "Don't worry, I'll help you get out of the ladder. I'll help you get out."\newline
So, Lucy and the ranger got out of the ladder and started to climb up the ladder. When they reached the bottom, Lucy was so happy. She thanked the ranger for helping her and said, "I'm glad I could help."\newline
The ranger smiled and said, "You're welcome, Lucy. I'm glad I could help." & Grammar: 5/10\newline
Creativity: 6/10\newline
Consistency: 3/10 \\
\midrule
8.3M \newline 8 layers & Lucy! What are you doing up there?"\newline
Lucy replied, "I'm trying to reach the top of the ladder, but I'm too small."\newline
The park ranger smiled and said, "Don't worry, I can help you. I can help you."\newline
The park ranger was very kind and said, "I can help you reach the top of the ladder if you want."\newline
Lucy was so excited and said, "Yes please!"\newline
The park ranger helped Lucy reach the top of the ladder and Lucy was so happy. She thanked the park ranger and said, "Thank you for helping me!"\newline
The park ranger smiled and said, "You're welcome, Lucy. I'm glad I could help." & 
Grammar: 7/10\newline
Creativity: 5/10\newline
Consistency: 5/10 \\
\midrule
28M \newline 8 layers & What are you doing up there, little one?"\newline
Lucy replied, "I wanted to see what was on top of the ladder, but it fell and I'm stuck!"\newline
The park ranger smiled and said, "Don't worry, I'll help you down." He carefully lifted Lucy off the ladder and put her safely back on the ground. \newline
Lucy thanked the park ranger and said, "I'm glad you were here to help me." \newline
The park ranger smiled and said, "It was my pleasure. I'm glad I could help." \newline
Lucy smiled and ran off to explore the rest of the park. & 
Grammar: 9/10\newline
Creativity: 6/10\newline
Consistency: 9/10 \\
\midrule
21M \newline 1 layer & Don't worry, Lucy! I'm here to rescue you!"\newline
Lucy quickly climbed up the ladder and rescued Lucy. She was so happy to be safe and thanked the park ranger for rescuing her.\newline
The park ranger smiled and said, "You're welcome, Lucy. I'm glad I could help."\newline
The park ranger smiled back and said, "You're welcome, Lucy. I'm glad I could help." & Grammar: 8/10\newline
Creativity: 3/10\newline
Consistency: 7/10 \\
\midrule
33M \newline 2 layers & What are you doing up there, Lucy?"\newline
Lucy replied, "I wanted to see what was on top of the ladder."\newline
The park ranger smiled and said, "Well, it looks like you have a lot of fun. But remember, it's important to be careful when exploring new places."\newline
Lucy thanked the park ranger and carefully climbed back down the ladder. From then on, she always remembered to be careful when exploring new places. & Grammar: 7/10\newline
Creativity: 6/10\newline
Consistency: 4/10  \\
\midrule
GPT2-XL & Lucy, if you want to come up, I'll take you up and there's a ladder that'll help you get down." \newline
It was a very kind and simple act, and after trying many doors and ladders, she finally managed to climb up the ladder. However, since the ladder was very heavy and couldn't be used to jump down off, she climbed up by making the ladder lean against the wall. \newline Her climbing was slow. First, she had to climb to a nearby rock, then she had to crawl to the cliff and then herassador and back. Finally, she managed to reach the top.& \\
\bottomrule
\end{tabularx}
\caption{Completion of different models - Example 1} \label{tab:story1}
\end{figure}

\newcolumntype{b}{X}
\newcolumntype{s}{>{\hsize=.3\hsize}X}

\begin{figure}[h!] 
\centering
\scriptsize
\begin{tabularx}{\textwidth}{p{1 cm}p{13 cm}p{2.5cm}}
\toprule
\textbf{Model} & \textbf{Completion} & \textbf{Scores} \\
\midrule
Prompt & Once upon a time there was a pumpkin. It was a very special pumpkin, it could speak. It was sad because it couldn't move. Every day, it would say & \\
\midrule
1M \newline 8 layers & hello to the pumpkin.\newline
One day, a little girl came to the pumpkin and saw the pumpkin. She was so happy and thanked the pumpkin. The pumpkin smiled and said, "Thank you, thank you!"\newline
The little girl and the pumpkin became good friends. They played together every day and had lots of fun. And they lived happily ever after. & Grammar: 4/10\newline
Creativity: 4/10\newline
Consistency: 3/10 \\
\midrule
2.5M \newline 8 layers & hello to the pumpkin.\newline
One day, a little girl came to the pumpkin. She was only three years old, but she was very curious. She asked the pumpkin what it was doing.\newline
The pumpkin said, "I want to speak to you!"\newline
The little girl said, "I want to speak to you!"\newline
The pumpkin said, "I want to speak to you!"\newline
The little girl said, "I want to talk to you!" ... & 
Grammar: 5/10\newline
Creativity: 4/10\newline
Consistency: 6/10 \\
\midrule
8.3M \newline 8 layers & "I want to move!"\newline
One day, a little girl saw the pumpkin. She was only three years old. She smiled and said, "Hello, pumpkin!"\newline
The pumpkin was so happy. It said, "Hello, little girl!"\newline
The little girl smiled and said, "Can I help you move?"\newline
The pumpkin said, "Yes, please!"\newline
So, the little girl and the pumpkin moved together. They moved around the garden, around the trees, and around the flowers. They had so much fun!\newline
When they were done, the pumpkin said, "Thank you, little girl. You are so kind."\newline
The little girl smiled and said, "You're welcome, pumpkin. I'm glad I could help."\newline
The pumpkin smiled and said, "Me too!"\newline
And they both lived happily ever after. & 
Grammar: 8/10\newline
Creativity: 7/10\newline
Consistency: 8/10 \\
\midrule
28M \newline 8 layers & 'hello' to everyone it met.\newline
One day, a little girl heard the pumpkin speaking. She was only three years old, but she was very brave. She walked up to the pumpkin and said, "Hello, pumpkin. Why are you so sad?"\newline
The pumpkin looked up at the little girl and said, "I am sad because I can't move."\newline
The little girl smiled and said, "Don't worry, pumpkin. I will help you move."\newline
The little girl put her hands on the pumpkin's back and said, "Come with me, pumpkin. I will take you to a place where you can be happy."\newline
The pumpkin smiled and said, "Thank you, little girl. I will be happy now."\newline
The little girl and the pumpkin walked together until they reached the place where the pumpkin could be happy. The pumpkin said goodbye to the little girl and thanked her for her help.\newline
The little girl smiled and said, "You're welcome, pumpkin. I'm glad I could help." & 
Grammar: 9/10\newline
Creativity: 7/10\newline
Consistency: 9/10 \\
\midrule
28M \newline 8 layers \newline \newline
temper-\newline ature 0.8 & hi, but no one ever answered.

One day, the pumpkin heard a voice. It was soft and gentle, and it said "Hello". The voice said "Hello!" The pumpkin was so happy!

The voice belonged to a very friendly bird. The bird flew above the pumpkin and said "I can help you move". The bird said "I can be your friend".

So the bird flew around the pumpkin and sang a song. The song was so nice and joyful that the pumpkin felt like it could move again.

The bird and the pumpkin became best friends and they had lots of fun together. Whenever the bird came, it would always be there to help the pumpkin. The end.
& 
Grammar: 7/10\newline
Creativity: 6/10\newline
Consistency: 9/10 \\
\midrule
21M \newline 1 layer & "Hello, pumpkin!"\newline
One day, a little girl came by and saw the pumpkin. She smiled and said, "Hello, pumpkin!" The pumpkin smiled back and said, "Hello, little girl!" The pumpkin was so happy to speak with the pumpkin. They talked and laughed together.\newline
From that day on, the pumpkin and the little girl were the best of friends. They would speak to each other every day, and the pumpkin would speak and laugh together. And they lived happily ever after. & Grammar: 7/10\newline
Creativity: 3/10\newline
Consistency: 3/10 \\
\midrule
33M \newline 2 layers & hello to all the other pumpkins in the garden.\newline
One day, a little girl came to the garden. She saw the pumpkin and said, "Hello, pumpkin! Why are you so sad?"\newline
The pumpkin replied, "I am sad because I cannot move. I want to stay here with my friends."\newline
The little girl smiled and said, "Don't worry, pumpkin. I will help you." She picked up the pumpkin and took it to her house.\newline
The little girl put the pumpkin on a table and said, "Now you can stay here with your friends."\newline
The pumpkin was so happy. It said, "Thank you, little girl. You are so kind."\newline
The little girl smiled and said, "You're welcome, pumpkin. I'm glad I could help."\newline
The little girl stayed with the pumpkin until it was time for her to go home. The pumpkin said, "Goodbye, little girl. I will miss you."\newline
The little girl said, "Goodbye, pumpkin. I will miss you." & Grammar: 9/10\newline
Creativity: 6/10\newline
Consistency: 8/10 \\
\midrule
GPT2-XL & Honey I gotta go to Paris.'" (He was referring to himself.) 'I've got friends'. "The way I look at my kids isn't that I'm the only one with friends. I've got friends, of course, and I have a wonderful relationship with them at home and my family over there. But I have a relationship with them with them on the field as well." "There is nothing but respect for the people we beat because they have been there before." "I have a very good relationship with [Drew] Stanton. I love Drew. I really do.& \\

\bottomrule
\end{tabularx}
\caption{Completion of different models with scores - Example 2} \label{tab:story2}
\end{figure}

\begin{figure}[h!] 
\centering
\scriptsize
\begin{tabularx}{\textwidth}{p{1 cm}p{13 cm}p{2.5cm}}
\toprule
\textbf{Model} & \textbf{Completion} & \textbf{Scores} \\
\midrule
Story beginning (prompt) & Once upon a time, there lived a black cat. The cat belonged to a little girl called Katie. Every day, Katie would take her cat for a walk in the park. \newline
One day, as Katie and her cat were walking around, they saw a mean looking man. He said he wanted to take the cat, to which she replied "This cat belongs & \\
\midrule
1M \newline 8 layers & to me". \newline
Katie said "No, I don't want to get hurt." \newline
Katie said "No, I want to be your friend." \newline
Katie said "No, I want to be your friend."... & Grammar: 2/10\newline
Creativity: 1/10\newline
Consistency: 1/10 \\
\midrule
2.5M \newline 8 layers & to me". \newline
Katie was very scared. She didn't know what to do. Then, she had an idea. She said to the man, "Please don't take my cat away". \newline
The man said, "No, I don't want you to take my cat away. I don't want you to get hurt". \newline
Katie was very sad. She didn't understand why the man was so mean. She said, "I'm sorry, I won't take your cat away". \newline
The man smiled and said, "It's okay, but you have to be more careful next time". \newline
Katie nodded and said, "Okay, I'll be more careful next time". \newline
The end. & Grammar: 5/10\newline
Creativity: 4/10\newline
Consistency: 5/10 \\
\midrule
8.3M \newline 8 layers & to me!". \newline
Katie was scared, but she didn't want to give up her cat. She said to the man, "No, this cat belongs to me!". \newline
The man didn't listen, and he grabbed the cat and ran away with it. Katie was very sad, but she knew she had done something wrong. \newline
The next day, Katie went back to the park and saw the man again. She asked him why he had taken her cat away. The man said, "I'm sorry, I didn't know it belonged to you." \newline
Katie was so happy to have her cat back. She gave him a big hug and said, "It's okay, I forgive you." \newline
From that day on, Katie and her cat were the best of friends. & Grammar: 7/10\newline
Creativity: 3/10\newline
Consistency: 6/10 \\
\midrule
28M \newline 8 layers & to me". \newline
Katie was scared, so she said to the man "Please don't take my cat". But the man didn't listen. He grabbed the cat and ran away. \newline
Katie started to cry, but the man didn't care. He took the cat away and Katie never saw her cat again. \newline
The end. & Grammar: 8/10\newline
Creativity: 5/10\newline
Consistency: 8/10 \\
\midrule
21M \newline 1 layer & to me!". \newline
The man said, "No, it belongs to me. I must give it back." \newline
Katie thought for a moment and then said, "I know! I will give it back to the cat." \newline
The man smiled and said, "That's very kind of you. Here, take it back." \newline
Katie thanked the man and took the cat home. She was so happy to have her own black cat back. & Grammar: 5/10\newline
Creativity: 3/10\newline
Consistency: 3/10 \\
\midrule
33M \newline 2 layers & to me!" \newline
Katie was scared, but she knew she had to protect the cat. So, she bravely stepped forward and said "No! This cat belongs to me!" \newline
The man was surprised by her courage and stepped forward. He said "I'm sorry, I didn't know it was yours. I won't take it again." \newline
Katie smiled and said "It's okay. I'm just glad you're safe."\newline
From that day on, Katie and her black cat were the best of friends. & Grammar: 9/10 \newline
Creativity: 6/10 \newline
Consistency: 8/10 \\
\midrule
GPT2-XL & to me, and I will not give it to you." The man then said "I will take you to court." Katie replied "I will not go to court, I will not give you my cat." The man then said "I will take you to jail." Katie replied "I will not go to jail, I will not give you my cat." The man then said "I will take you to the hospital." Katie replied "I will not go to the hospital, I will not give you my cat." The man then said "I will take you to the morgue." Katie replied "I will not go to the morgue, I will not give you my cat." The man then said "I will take you to the crematorium." Katie replied "I will not go to the crematorium, I will not give you my cat." & \\
\bottomrule
\end{tabularx}
\caption{Completion of different models with scores - Example 3} \label{tab:story3}
\end{figure}

\begin{figure}
\scriptsize
\begin{tabular}{p{3cm}p{1.4cm}p{1.4cm}p{1.4cm}p{1.4cm}p{1.4cm}p{1.4cm}p{1.4cm}p{1.4cm}}
\toprule
\textbf{Prompt} & \textbf{1M \newline 8 layers} & \textbf{2.5M \newline 8 layers} & \textbf{8.3M \newline 8 layers} & \textbf{28M \newline 8 layers} & \textbf{33M \newline 4 layers} & \textbf{21M \newline 1 layer} & \textbf{33M \newline 2 layers} & \textbf{GPT2-XL (1.5B)} \\
\midrule
Alice was so tired when she got back home so she went & \cellcolor{red!50} home. & \cellcolor{yellow!50} to bed. & \cellcolor{yellow!50} to bed. & \cellcolor{green!50} straight to bed. & \cellcolor{green!50} straight to bed. & \cellcolor{yellow!50} to bed with a big smile on her face. & \cellcolor{green!50} straight to bed. & \cellcolor{red!50} outside\\
\midrule
Jack and Lily saw a rainbow after a rainy day. They were amazed by the colors. Jack said, "Look, Lily. A rainbow has & \cellcolor{red!50} a rainbow! & \cellcolor{yellow!50} so many colors. & \cellcolor{yellow!50} many colors! & \cellcolor{yellow!50} appeared. It's so pretty. & \cellcolor{green!50} red, orange, yellow, green, blue, and purple! & \cellcolor{yellow!50} many colors. Do you like rainbows? & \cellcolor{yellow!50} many colors. & \cellcolor{yellow!50} appeared! \\
\midrule
Jack and Lily liked to watch the moon at night. They noticed that the moon changed its shape every night. Sometimes the moon was big and round, and sometimes it was & \cellcolor{red!50} different. & \cellcolor{red!50} like the moon. & \cellcolor{yellow!50} small. & \cellcolor{red!50} small and round. & \cellcolor{green!50} small and thin. & \cellcolor{yellow!50} small. & \cellcolor{green!50} small and thin. & \cellcolor{yellow!50} small and round. \\
\midrule
Jack wanted to read a book, so he went to & \cellcolor{red!50} his mom's house. & \cellcolor{red!50} his mom and said, "Mommy, can I read this book?" & \cellcolor{yellow!50} the bookshelf. & \cellcolor{green!50} the library. & \cellcolor{green!50} the library. & \cellcolor{yellow!50} his mom and said, "Mom, I want to read this book!" & \cellcolor{yellow!50} the bookshelf & \cellcolor{yellow!50} the bookshelf \\
\midrule
"Can cows fly?", Alice asked her mother. & \cellcolor{red!50} Her mother smiled and said, "Yes, let's go!" & \cellcolor{red!50} "Yes, of course," her mother said. & \cellcolor{red!50} "Yes, sweetheart," her mother replied. & \cellcolor{green!50} Her mother smiled and said, "No, sweetheart, cows can't fly." & \cellcolor{green!50} "No, they can't fly," her mother said. & \cellcolor{red!50} "Yes!", her mother replied. Alice and her mother went to the barn. & \cellcolor{green!50} "No, sweetheart," her mother said. "They can't fly." & \cellcolor{red!50} "I suppose not, but the cows are allowed to fly", her mother replied.  \\
\midrule
"What do birds like to eat?", Tom asked his mother. & \cellcolor{red!50} His mother smiled and said, "That sounds like fun!" & \cellcolor{yellow!50} "I don't know," said his mother. "Let's go to the park." & \cellcolor{yellow!50} "I don't know," his mother replied. & \cellcolor{red!50} His mother smiled and said, "They like to fly in the sky, Tom. That's why they have wings." & \cellcolor{green!50} His mother smiled and said, "They like to eat worms and bugs." & \cellcolor{red!50} Tom thought for a moment and then said, "Why don't you eat something healthy?" & \cellcolor{yellow!50} "I don't know" said his mother, let's go find out. & \cellcolor{yellow!50} "I dunno. Some things that look like animals. But he'll tell you the whole story" \\
\midrule
"What language do they speak in France?", Tom asked his mother & \cellcolor{red!50} His mother smiled and said, "That sounds like fun, Tom. Let's go to the park ". & \cellcolor{red!50} Tom thought for a moment. Then he had an idea. He said to his mother, "Let's go to France!" & \cellcolor{yellow!50} His mother smiled and said, "It's called French. It's called French." & \cellcolor{red!50} His mother smiled and said, "They speak English." & \cellcolor{green!50} "They speak French," his mother replied. & \cellcolor{red!50} "It's called Spanish," his mother replied. & \cellcolor{yellow!50} "It's a language that we speak to each other," his mother replied. & \cellcolor{green!50} "In France they speak French", she replied\\
\midrule
If I throw a ball up in the air, eventually it will & \cellcolor{red!50} be time to go home. & \cellcolor{red!50} be fun!" & \cellcolor{green!50} come down." & \cellcolor{green!50} come down." & \cellcolor{green!50} come down.” & \cellcolor{red!50} come true! & \cellcolor{green!50} come back down. & \cellcolor{green!50} come down \\
\midrule
It was winter and cold outside so his mother told him, "You should & \cellcolor{yellow!50} be careful and stay safe. & \cellcolor{yellow!50} go outside and play in the snow. & \cellcolor{yellow!50} go outside and play in the snow. & \cellcolor{green!50} wear your warm coat, it will keep you warm. & \cellcolor{green!50} wear your warm coat so you don't get cold. & \cellcolor{green!50} wear warm clothes so you don't get sick. & \cellcolor{green!50} stay inside where it's warm and cozy. & \cellcolor{red!50} You should ask a female friend to marry you. \\
\bottomrule
\end{tabular}
\caption{Performance of different models on factual prompts} \label{fig:factual}
\end{figure}

\begin{figure}
\scriptsize
\begin{tabular}
{p{3cm}p{1.4cm}p{1.4cm}p{1.4cm}p{1.4cm}p{1.4cm}p{1.4cm}p{1.4cm}p{1.4cm}}
\toprule
\textbf{Prompt} & \textbf{1M \newline 8 layers} & \textbf{2.5M \newline 8 layers} & \textbf{8.3M \newline 8 layers} & \textbf{28M \newline 8 layers} & \textbf{33M \newline 4 layers} & \textbf{21M \newline 1 layer} & \textbf{33M \newline 2 layers} & \textbf{GPT2-XL (1.5B)} \\
\midrule
Lily likes cats and dogs. She asked her mom for a dog and her mom said no, so instead she asked &  \cellcolor{red!50}  her mom.  &  \cellcolor{red!50}  her mom if she could have a dog. &  \cellcolor{red!50}  her dad for a dog.  &  \cellcolor{green!50}  her dad for a cat.  &  \cellcolor{green!50}  her dad for a cat.  &  \cellcolor{red!50}  her mom again.  &  \cellcolor{yellow!50}  her dad for a dog and her mom said yes. & \cellcolor{red!50} her dad. They brought her a Chihuahua, and she took to them immediately. \\ 
\midrule
Jack told Mary, "If you give me your banana, I'll give you my apple". Mary gave Jack her Banana so &  \cellcolor{red!50}  she gave him a big hug. &  \cellcolor{red!50}  she gave him a big hug.   &  \cellcolor{yellow!50}  he could have it.  &  \cellcolor{red!50}  he gave her the banana. &  \cellcolor{green!50}  he could give her the apple.  &  \cellcolor{yellow!50}  he could have it. &  \cellcolor{yellow!50}   he could eat the banana. & \cellcolor{yellow!50} that he was free to eat it.\\ 
\midrule
On weekends Jack went to visit his grandmother whereas on weekdays he would go to school. Last weekend, when Jack was on his way to &  \cellcolor{red!50}  school &  \cellcolor{red!50}  school &  \cellcolor{red!50}  school &  \cellcolor{red!50}  school &  \cellcolor{green!50}  his grandmother's house &  \cellcolor{green!50}  Grandma's house &  \cellcolor{red!50}  school &  \cellcolor{red!50}  school \\ 
\midrule
Lily and Ben were having an argument. Ben said that cake is much better than ice cream and Lily said that &  \cellcolor{red!50}  it was too expensive &  \cellcolor{red!50}  they should share and be kind to each other &  \cellcolor{yellow!50}  cake is yucky   &  \cellcolor{yellow!50}  cake was boring &  \cellcolor{green!50}  ice cream is better than cake &  \cellcolor{red!50}  was not fair &  \cellcolor{red!50}  ice cream is better than ice cream. &  \cellcolor{red!50} the ice cream was good enough \\ 
\midrule
Lily and Ben are having an argument. They are trying to decide between the park and the swimming pool. Ben says, "I want to go to the park". Lily says &  \cellcolor{red!50} , "No, Ben, I want to play with you. &  \cellcolor{green!50} , "No, I want to go to the pool. It &  \cellcolor{green!50} , "No, I want to go to the pool $\langle \dots \rangle "$ &  \cellcolor{green!50} , "No, I want to go to the pool."  &  \cellcolor{green!50} , "No, I want to go to the pool".  &  \cellcolor{red!50} , "No, I want to go to the pool. &  \cellcolor{red!50} , "No, I want to go to the park". & \cellcolor{yellow!50} "Just for once, let's not make a bigoted choice"\\ 
\midrule
Jack's mother was not home, and his father was at home. When Jack came home, he said hello to &  \cellcolor{red!50}  Jack and said, "Hi Jack! I'm so happy to meet you!"&  \cellcolor{green!50}  his father &  \cellcolor{green!50}  his father &  \cellcolor{green!50}  his father &  \cellcolor{green!50}  his mother, but she didn't answer.  Jack was confused. &  \cellcolor{red!50}  his mother.  "Where did you get this, Jack?" his mother asked.  &  \cellcolor{green!50}  his father & \cellcolor{red!50} his mother \\ 
\midrule
Lily doesn't like swimming. When her father wants to take her to the swimming pool, she says &  \cellcolor{red!50} , "Lily, you can swim in the pool."  &  \cellcolor{red!50}  yes.  Lily and her father went to the pool.  &  \cellcolor{green!50} , "Daddy, I don't want to go to the pool". &  \cellcolor{green!50}  no. She wants to stay at home and play with her toys. &  \cellcolor{green!50} , "No, I don't want to go. I want to stay here."   &  \cellcolor{red!50} , "No, I don't want to swim. I want to stay in the water." &  \cellcolor{green!50}  no.  & \cellcolor{green!50} she isn't going until he has convinced her to go. \\ 
\midrule
Both Ben and Lily wanted cake. Father said that there was only one piece of cake left. They &  \cellcolor{red!50}  said that they had to wait for the cake. &  \cellcolor{red!50}  went to the store and bought the cake. &  \cellcolor{green!50}  did not want to share. They started to fight over the cake. &  \cellcolor{green!50}  both reached for it at the same time.  Father said, "If you don't share the cake, I will have to punish you." &  \cellcolor{green!50}  started to fight over the cake. They pulled and pushed and shouted. &  \cellcolor{yellow!50}  started to argue and fight over the cake.  &  \cellcolor{red!50}  both wanted it.  & \cellcolor{yellow!50} couldn't have any of the cake \\ 
\midrule
Ben went to visit Lily in her house, but she was not at home. Ben knocked on the door, &  \cellcolor{red!50}  but it was too late. Lily was sad and angry. &  \cellcolor{red!50}  and Lily's mom opened it.  "Ben, what are you doing?" Lily asked. &  \cellcolor{green!50}  but no one answered. He knocked again, but still no one answered. &  \cellcolor{green!50}  but no one answered. He tried to open the door, but it was locked. &  \cellcolor{green!50}  but no one answered. He knocked again and again, but still no one came. &  \cellcolor{yellow!50}  but no one answered. Lily felt sad and lonely. &  \cellcolor{green!50} but Lily did not answer. & \cellcolor{green!50} but no one came to answer. \\
\bottomrule
\end{tabular}
\caption{Performance of different models on reasoning prompts} \label{fig:reasoning}
\end{figure}

\begin{figure}
\scriptsize
\begin{tabular}{p{3cm}p{1.4cm}p{1.4cm}p{1.4cm}p{1.4cm}p{1.4cm}p{1.4cm}p{1.4cm}p{1.4cm}}
\toprule
\textbf{Prompt} & \textbf{1M \newline 8 layers} & \textbf{2.5M \newline 8 layers} & \textbf{8.3M \newline 8 layers} & \textbf{28M \newline 8 layers} & \textbf{33M \newline 4 layers} & \textbf{21M \newline 1 layer} & \textbf{33M \newline 2 layers} & \textbf{GPT2-XL (1.5B)} \\
\midrule
"Hi Jane, have you seen Alice? I can't find her anywhere", said Jack. & \cellcolor{green!50} Jane looked around, but she couldn't find her.  & \cellcolor{green!50} Jane smiled and said, "Yes, I can help you. Let's go!" & \cellcolor{green!50} "I can't find her anywhere," said Jane. & \cellcolor{green!50} Jane smiled and said, "Don't worry, I'll help you find her. Let's look together". & \cellcolor{green!50} Jane smiled and said, "Don't worry Jack, I'll help you find her". Jack and Jane looked around the park, but they couldn't find Alice. & \cellcolor{red!50} Alice looked around the park, but she couldn't find Jack anywhere. & \cellcolor{green!50} Jane smiled and said, "Yes, I think I saw her in the park. Let's go look together!" & \cellcolor{green!50} "I think she's up in the trees, by the lake," she replied. \\
\midrule
Max had two dogs. One was white and the other was black. Max walked up the street and saw a kid with a dog. He told the kid, "I see you have a Brown dog. I also have & \cellcolor{yellow!50} a pet dog. & \cellcolor{red!50} a leash. & \cellcolor{yellow!50} a dog. & \cellcolor{green!50} two dogs. & \cellcolor{green!50} two dogs. & \cellcolor{red!50} a red ball. & \cellcolor{yellow!50} a dog. & \cellcolor{red!50} a Brown dog \\
\midrule
Anne had a piece of candy in her left pocket and a piece of chocolate in her right pocket. Anne's mom asked her, "Anne, what is that you have in your left pocket?" & \cellcolor{red!50} Anne smiled and said, "I want to give it back!" & \cellcolor{red!50} Anne said, "I'm sorry, Mommy. I didn't know it was yours." & \cellcolor{green!50} Anne smiled and said, "It's a piece of candy, Mommy." & \cellcolor{green!50} Anne replied, "It's a piece of candy, Mommy!" & \cellcolor{green!50} Anne replied, "It's a piece of candy, Mommy. It's so yummy!" & \cellcolor{red!50} Anne shook her head and said, "No, I found it in my pocket." & \cellcolor{green!50} Anne smiled and said, "It's a piece of candy. I want to eat it!" & \cellcolor{yellow!50} The girl responded, "I have pieces of candy." \\
\midrule
Alice had both an apple and a carrot in her bag. She took the apple out of the bag and gave it to Jack. She reached into the bag again and took & \cellcolor{red!50} it home. & \cellcolor{red!50} a big bite of the apple. & \cellcolor{green!50} out the carrot. & \cellcolor{green!50} out the carrot. & \cellcolor{green!50} out the carrot. & \cellcolor{red!50} out the apple. & \cellcolor{yellow!50} the carrot and the carrot and the carrot. & \cellcolor{green!50} the carrot \\
\midrule
Alice and Jack walked up the street and met a girl in a red dress. The girl said to them, "Hi, I'm Jane. What are your names?" & \cellcolor{red!50} Jane asked, "Do you want to play with me?" & \cellcolor{red!50} Alice smiled and said, "My name is Daisy. What's your name?" & \cellcolor{yellow!50} Alice said, "I'm Alice. What's your name?" & \cellcolor{green!50} Alice smiled and said, "I'm Alice and this is Jack." & \cellcolor{green!50} Alice said, "I'm Alice and this is Jack." & \cellcolor{red!50} Jack smiled and said, "I'm Jane. Nice to meet you!" & \cellcolor{yellow!50} Jack said, "I'm Jack. I'm three years old." & \cellcolor{red!50} Jane said, "Jack and Kate"\\
\midrule
Diva was hungry, and wanted to bake a cake, but she didn't have any sugar at home, so she decided to go ask around. She started walking and met a squirrel. She asked the squirrel, "Would you happen & \cellcolor{red!50} now?" & \cellcolor{red!50} to me?" & \cellcolor{red!50} to bake a cake?" & \cellcolor{yellow!50} to help me bake a cake?" & \cellcolor{green!50} to have some sugar?" & \cellcolor{red!50} to my cake?" & \cellcolor{red!50} to bake a cake with me?" & \cellcolor{green!50} to have some sugar? \\
\bottomrule
\end{tabular}
\caption{Performance of different models on context-tracking prompts.} \label{fig:consistency}
\end{figure}

\subsection{Instruction-following examples and out-of-distribution generation}
Table \ref{tab:story4} provides an example of the generation of different models trained on the \TNI~dataset, together with the evaluation scores given by GPT-4. As the model size increases, we see an improvement both its ability to follow instructions and to generate a coherent plot.

This dataset also enables us to test whether our models have a reasonable \textbf{out of distribution} performance. Recall that in each entry of \TNI, the instructions are created as a (random) combination of possible types of instructions (words to use, summary, prescribed sentence, features). We created another variant of the \TNI~(called \TNI-OOD) where we disallowed one specific combination of instruction-types: The dataset does not contain any entry where the instruction combines both the summary of the story and the words that the story needs to use (we chose this particular combination because in a sense, it is the most restrictive one). We then tested whether models trained on this variant would be able to produce stories that follow these two types of instructions \textbf{combined}. An example is provided in Figure \ref{fig:ood}, for a model with 33M parameters. We see that, perhaps somewhat surprisingly, the model is able to follow these two types of instructions simultaneously even if it has never been trained on such a task.

\begin{figure}[h!] 
\centering
\scriptsize
\begin{tabularx}{\textwidth}{sbs}
\toprule
\textbf{Prompt} & \textbf{Completion} & \textbf{Scores} \\
\midrule
Instructions (prompt) & Write a story containing the words: dive, job, sorry. \newline
Story summary: Bob the big fish finds a shiny rock while searching for food for his friends, but when he tells them about it, they are excited to play with it instead of being sad about not having food. & \\
\midrule
1M \newline 8 layers & Once upon a time, there was a big fish named Bob. Bob loved to swim and play with his friends. One day, Bob saw a shiny rock on the ground. He wanted to play with it, but it was too big for him.\newline
Bob asked his friends if they could play with the rock. His friends said yes, and they all played with the rock all day. They had so much fun playing together.\newline
But then, something unexpected happened. A big wind came and blew the rock away. Bob was sad and didn't know what to do. He looked around and saw that the rock was gone! He was sad and didn't know what to do. & Grammar: 4/10\newline
Creativity: 3/10\newline
Plot: 4/10\newline
Consistency: 6/10 \\
\midrule
2.5M \newline 8 layers & Once upon a time, there was a big fish named Bob. Bob loved to swim in the big blue sea. One day, Bob found a shiny rock. He wanted to show it to his friends.\newline
Bob swam to his friend, the big fish named Sam. Bob told Sam about the shiny rock. Sam said, "That's a big rock, Bob! Let's play with it!" So, Bob and Sam played with the shiny rock all day.\newline
At the end of the day, Bob and Sam were very tired. They went back to their homes. Bob told Sam about the shiny rock and the shiny rock. Sam said, "Thank you, Bob! You are a good friend." Bob smiled and said, "You're welcome, Sam!" & Grammar: 6/10\newline
Creativity: 4/10\newline
Plot: 5/10\newline
Consistency: 7/10 \\
\midrule
8.3M \newline 8 layers & Once upon a time, there was a big fish named Bob. Bob loved to dive deep in the water. One day, while diving, he found a shiny rock. He was very happy and wanted to show it to his friends.\newline
Bob swam to his friends and said, "Look what I found! It's a shiny rock!" His friends looked at the rock and said, "Wow! That's a nice rock, Bob!" They all wanted to play with the rock, but they were too scared to play with it.\newline
Bob felt sorry for the rock. He said, "I'm sorry, rock. I didn't mean to scare you. Let's play with it together!" So, Bob and his friends played with the shiny rock all day long. They had lots of fun and were very happy. & Grammar: 7/10\newline
Creativity: 6/10\newline
Plot: 6/10\newline
Consistency: 8/10 \\
\midrule
28M \newline 8 layers & Once upon a time, there was a big fish named Bob. Bob had a job. His job was to dive deep in the water to find food for his friends. He liked his job very much.\newline
One day, while Bob was diving, he found a big, shiny rock. He thought it was very pretty. He wanted to show it to his friends. But when he dove down, he saw that his friends were not happy. They were sad because they did not have food.\newline
Bob felt sorry for his friends. He said, "I am sorry, friends. I did not find food for you." His friends smiled and said, "It is okay, Bob. We can play with the rock instead." So, Bob and his friends played with the shiny rock and had lots of fun. & Grammar: 7/10\newline
Creativity: 6/10\newline
Plot: 7/10\newline
Consistency: 9/10 \\
\midrule
33M \newline 4 layers & Once upon a time, there was a big fish named Bob. Bob had a job. His job was to dive deep in the water to find food for his friends. One day, while diving, he saw something shiny.\newline
Bob swam closer to see what it was. It was a big, shiny rock. He wanted to show it to his friends. But when he got to his friends, they were not happy. They did not want to play with the shiny rock.\newline
Bob felt sorry. He did not want to make his friends sad. So, he told them about the shiny rock. They all went to see it together. They were excited to play with the shiny rock.\newline
In the end, Bob's friends were happy. They played with the shiny rock and had lots of fun. Bob was glad he did his job and made his friends happy. And they all lived happily ever after. & Grammar: 8/10\newline
Creativity: 7/10\newline
Plot: 8/10\newline
Consistency: 9/10 \\
\midrule
21M \newline 1 layer & Once upon a time, there was a little fish named Bob. Bob loved to dive deep in the water. One day, while diving, he found a shiny rock. Bob was very happy and wanted to show it to his friends.\newline
Bob went to his friend, the fish, and said, "Look, I found a shiny rock!" His friend, the fish, said, "Wow, Bob! That's a very pretty rock. Can I see it?" Bob thought for a moment and said, "No, it's mine. I found it first."\newline
Bob was sad, but he didn't give up. He went to his friend, the fish, and said, "I found this shiny rock. It's mine now." His friend was very happy and said, "Thank you, Bob! You are a good friend."\newline
From that day on, Bob and his friend shared the shiny rock and played together every day. They were both very happy and had lots of fun. & Grammar: 7/10\newline
Creativity: 5/10\newline
Plot: 6/10\newline
Consistency: 4/10 \\
\midrule
33M \newline 2 layers & Once upon a time, there was a big fish named Bob. Bob had a job. His job was to dive deep in the water and find food for his friends.\newline
One day, while Bob was diving, he saw a shiny rock. He wanted to show his friends the shiny rock. But when he told his friends about the rock, they were sad. They did not have any food to eat.\newline
Bob felt sorry for not having food. He told his friends about the shiny rock. They were excited to play with it. They all went to the shiny rock and played together. They were happy again. & Grammar: 7/10\newline
Creativity: 6/10\newline
Plot: 7/10\newline
Consistency: 8/10 \\
\bottomrule
\end{tabularx}
\caption{Performance of different models on an instruction-following instance} \label{tab:story4}
\end{figure}

\begin{figure}
\fbox{
\parbox{\textwidth}
{
\textbf{Words}: come, road, sad \\
\textbf{Summary}: A bus becomes jealous of a shiny new car and undergoes an operation to go faster, becoming good friends with the car and making everyone in the town happy. \\
\textbf{Story}: \hl{Once upon a time, there was a big red bus. The bus loved to go fast on the road. One day, the bus saw a new car on the road. The car was shiny and new. The bus felt sad because it wanted to be shiny too. \\

The bus went to the car and said, "Can I come with you?" The car said, "Yes, you can come with me." So, the bus went with the car on the road. They went very fast together. The bus was so happy to be with the car. \\

But then, something unexpected happened. The car started to go faster and faster. The bus was scared. The car said, "Don't worry, I will help you." The car stopped and the bus was safe. The bus and the car became good friends. They went fast on the road together and made everyone in the town happy.} } }

    \caption{Out-of-distribution generations: stories generated by a model trained on a dataset where the "Words" and "Summary" parts do not appear together.}
    \label{fig:ood}
\end{figure}

\subsection{Diversity of the content generated by the model} \label{sec:diversity}

One of the main challenges of text generation is to produce diverse and creative texts that are not just repetitions or variations of existing texts. Our small models can generate coherent and fluent English text, but this would not be very impressive if they were simply copying or paraphrasing large portions of the dataset. Therefore, in this section, we aim to address this concern. We will provide several methods and metrics that show that the models can generate diverse texts that are not similar to any story in the dataset, and that they can adapt to different instructions and contexts.

To evaluate the diversity of the content generated by the models, we first need to define what we mean by memorization, and what kinds of memorization we want to avoid or detect. We classify three levels of memorization as follows:

\begin{itemize}
\item Exact memorization: This is the simplest and most obvious form of memorization, where the model simply copies an entire story or a large portion of it from the dataset, without changing anything. This can be easily detected by checking the similarity or the hash of the generated story with the stories in the dataset.
\item Simple template matching: This is a slightly more sophisticated form of memorization, where the model changes some names or entities in a story from the dataset, but keeps the rest of the story the same. For example, the model might change the names of characters, or the location of the story, but keep the plot and the events the same. This can be detected and prevented by measuring the overlap of words and n-grams between the generated story and the stories in the dataset.
\item Complex template matching: This is the most subtle and difficult form of memorization, where the model follows a more abstract pattern or structure from the dataset, keeping the general plot but changing the details and the specifics of the story. This is almost impossible to quantify, as it requires a deeper understanding and analysis of the content and the meaning of the stories, and how they relate to each other. 
\end{itemize}

We claim that our models are not doing exact memorization or simple template matching, as evidenced by the methods and metrics we use to evaluate the diversity of the content generated by the models. We rely on several approaches:

\begin{itemize}
\item Manual inspection: We generate completions for a range of human-constructed stories. We inspect the stories generated by the models and check that they are not copies or close modifications of the stories in the dataset.
\item Completion of training stories: We take stories from the training set, truncate them in the middle and generate alternative completions with our models. We then compare the completions with the original stories. We observe that the completions are typically very different from the original stories, and often introduce new characters, events, or twists. This is shown in Figure~\ref{fig:r1}.
\item Diversity of instructions: Recall that in the \TNI~dataset, we provide a set of instructions in the form of summaries or words contained in the stories, followed by the stories themselves. We can then change the instructions, verify that the combinations do not appear in the dataset and see how the models adapt to the new instructions. We find that the models can generate diverse stories that follow the instructions, even if they are novel or challenging, such as requiring the model to fit unlikely words into the story or adding features such as a plot twist or a bad ending.
\end{itemize}

\subsubsection{Quantitative measurement of similarity using Rouge score.}
We measure the diversity of the stories quantitatively using word and n-gram overlap. We inspect the overlap of words and n-grams between different stories generated by the models, and compare them with the overlap in the dataset. We find that the models' generations have a very low overlap with the dataset, indicating that they are not repeating the same words or phrases. We use the standard Rouge score, for the source text $T_1, T_2$ with $k$-gram $\mathcal{G}_k(T_1), \mathcal{G}_k(T_2)$ respectively, the rouge$k$ precision score is defined as:
$$R_{k, p}(T_1, T_2) = \frac{1}{|\mathcal{G}_k(T_1)|} \sum_{t \in \mathcal{G}_k(T_1)} 1_{t \in \mathcal{G}_k(T_2)}.$$

The Rouge$k$ precision score measures how many $k$-grams in $T_1$ is included in that of $T_2$. The final Rouge$k$ score (fmeasure) is given as:
$$R_{k}(T_1, T_2) =   \frac{2R_{k}(T_1, T_2) \times R_{k}(T_2, T_1) }{ R_{k}(T_1, T_2) + R_{k}(T_2, T_1)}.$$

\begin{figure}[h!]
\includegraphics[width=1\textwidth]{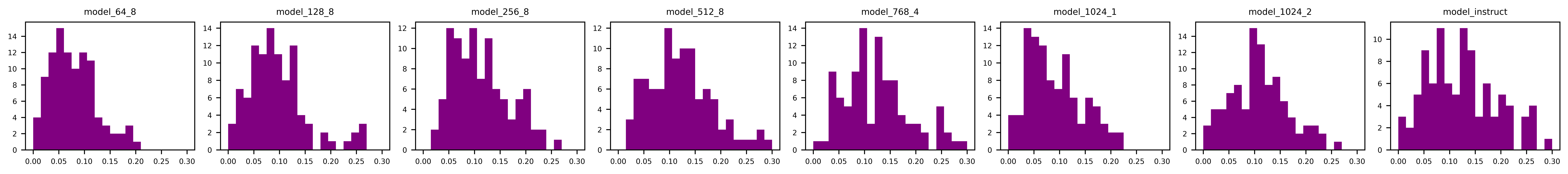}
\caption{Rogue2 (precision) score between the model's completion and the original story from the same beginnings (we select 100 from the training dataset). We can see that most of the completions that the models generate are very different from the ones in the training dataset (and also not subsampled versions of the original ones).} \label{fig:r1}
\end{figure}

\begin{figure}[h!]
\includegraphics[width=1\textwidth]{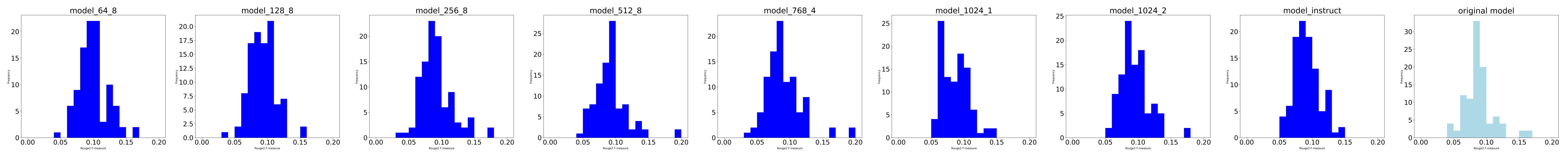}
\caption{Maximum Rouge2 score (fmeasure) similarity between the 100 generated stories for each model. Here original model means the ones generated by GPT-3.5.} \label{fig:r2}
\end{figure}

\begin{figure}[h!]
\includegraphics[width=1\textwidth]{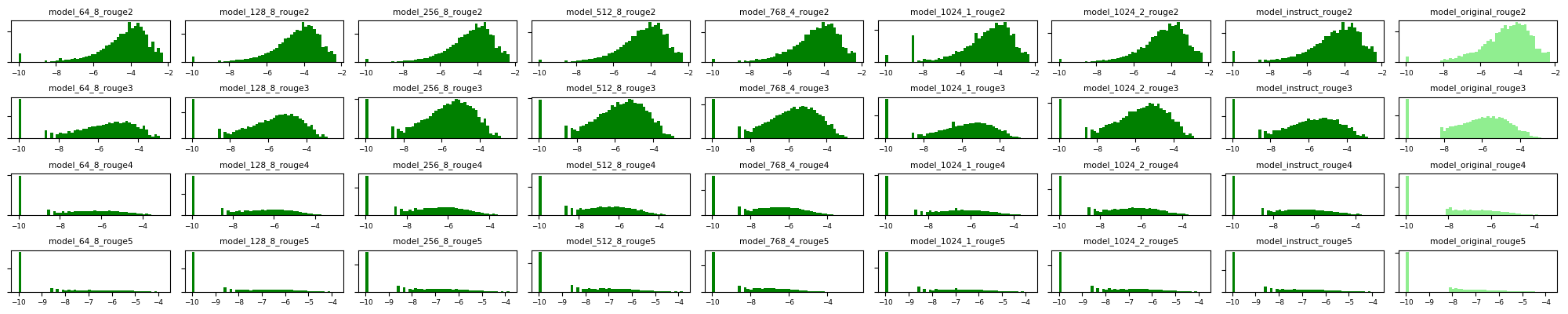}
\caption{Histogram plot of how many times (fraction) each k-gram in the models' generations also appears in the training data in log scale (base 10). $-10$ means it never appears. We can see that most of the 4, 5-grams in the models' generations do not even appear once in the entire training data.} \label{fig:r3}
\end{figure}

\begin{figure}[h!]
\begin{center}
\includegraphics[width=1\textwidth]
{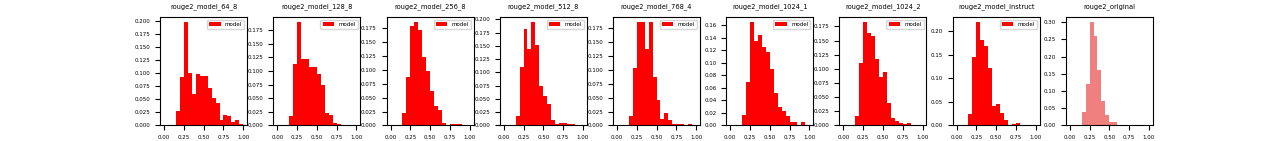}
\includegraphics[width=1\textwidth]
{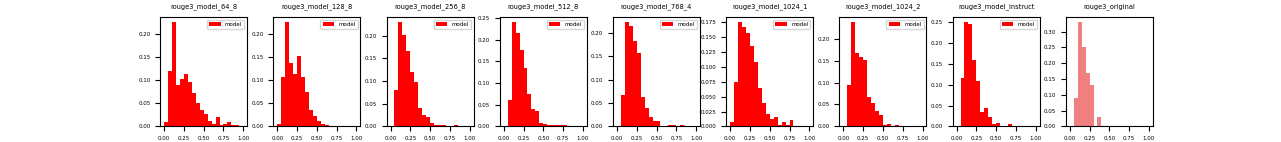}
\end{center}
\caption{Histogram plot for each generated story, the highest rougek score (precision) to the stories in the training dataset. We can see that the models' generations are not copying from any particular story in the training dataset.} \label{fig:r4}
\end{figure}

We perform the following experiment: We randomly pick $100$ stories from the training dataset, we cut each story in the middle, keeping roughly the first $40\%$, and use it as a prompt. We ask the model to generate a completion from each prompt. Let $T_1, T_2 ,\cdots, T_{100}$ be the generated completions and $T_1', T_2', \cdots, T_{100}'$ be the original completion, we measure: 
\begin{enumerate}
\item 
How much of the new generation is contained in the original story (Figure~\ref{fig:r1}), meaning:
$$s_i := R_{2, p} (T_i, T_i').$$
\item
How similar are the generated $100$ stories to each other (Figure~\ref{fig:r2}), meaning:
$$r_i  := \max_{j \not= i} R_2(T_i, T_j)$$
\item
To what extent are the k-grams in the generated story copied from the training dataset (Figure~\ref{fig:r3}). More precisely, we take $S$ as the entire training corpus, for each $r \in \mathcal{G}_k(\{T_i\}_{i \in [100]})$  we measure
$$g_{r}  := \frac{\sum_{q \in  \mathcal{G}_k(S) } 1_{g_{r} = q} }{|\sum_{q \in  \mathcal{G}_k(S) } |}$$

In other words, for each $k$-gram generated by the model, we measure the frequency that it appears in the original training dataset, where $g_{r} = 0$ means that the $k$-gram never appears in the training dataset.
\item 
How similar is the generated story to the closest point, in terms of Rouge precision score, in the entire dataset. Let $S_1, S_2, \cdots, S_m$ be all the stories in the training dataset, in Figure~\ref{fig:r4}, we compute 
$$h_i = \max_{j \in [m]} R_{2, p}(T_i, S_j)$$
\end{enumerate} 


For the sake of getting a more concrete impression about how different the model completions are from the original ending of the story and from other stories in the dataset, in Figure \ref{fig:closestpoint} we provide one example of the original story, the alternative completion by our model together with its closest point in the training dataset.

\begin{figure}[h!]
\scriptsize
\centering
\begin{tabular}{p{0.2\textwidth} p{0.7\textwidth}}
\toprule
Beginning (prompt) & Tim and Jen are playing with their toys in the living room. Tim has a laser that makes red lights and sounds. Jen has a doll that she can dress and comb. Tim likes to make his laser point at Jen's doll and make noises. Jen does not like that. She tells Tim to stop.

"Stop, Tim! You are hurting my doll! She does not like lasers!" Jen says.

"No, I am not hurting her. I am just playing. Lasers are fun!" Tim says.

Tim does not stop. He makes his laser point at Jen's doll more and more. Jen gets angry. She grabs her doll and runs to her room. She slams the door and locks it. Tim follows her and knocks on the door.

"Jen, open the door! I want to play with you!" Tim says.

"Go away, Tim! You are mean! I don't \\ \midrule

Original story continuation & want to play with you!" Jen says.\newline Tim feels sad. He does not understand why Jen is mad at him. He thinks lasers are fun. He goes back to the living room and lays on the couch. He plays with his laser by himself. He points it at the ceiling, the walls, the floor. He does not see the big vase on the shelf. He points his laser at the vase. The vase falls and breaks. The pieces fly everywhere. One piece hits Tim on the head.

(rest of story omitted)
\\
\midrule
Completion by model trained on \TN~(28M) & want to play with you or your laser!" Jen says.

Tim does not listen. He keeps pressing the red light on Jen's doll. He thinks Jen is being nosy and rude. He waits for Jen to say something. Then he starts to play with his laser by himself. He makes sparks and flashes with his laser. He laughs and runs around the room.

Jen watches Tim from her window. She feels sad. She wishes Tim would play with her. She picks up her doll and goes to the door. She shakes the door and tries to open it. But the door does not open. There is a lock on the doorknob.

Jen has an idea. She runs to her room and gets her scissors. She comes back to the living room and puts the scissors in the lock. She presses the buttons on the door. She hopes it will open.

(rest of completion omitted)
\\ \midrule
\TN~training data entry with maximum similarity to the model's completion (Rouge precision score = 0.29) & Lila and Tom are playing with their toys in the living room. Lila has a smooth doll with long hair and a pink dress. Tom has a horn that makes a loud noise when he blows it. Lila likes to comb her doll's hair and make her look pretty. Tom likes to make his horn sound and scare Lila.

"Tom, stop it!" Lila says. "Your horn is too loud. It hurts my ears."

"But it is fun!" Tom says. "Look, I can make it sound like a car, or a cow, or a lion!"

He blows his horn again and again, making different noises. Lila covers her ears and frowns. She does not like Tom's horn. She wants him to be quiet.

"Tom, please shut your horn!" Lila says. "I want to play with my doll. She does not like loud noises. She likes soft music and nice words."

(rest of story omitted)
 \\
\bottomrule
\end{tabular}
\caption{The closest point in the dataset to an alternative completion}
\label{fig:closestpoint}
\end{figure}

The above points towards several findings:
\begin{itemize}
\item 
When the model generates stories using a diverse set of prompts, it ends up with a diverse set of completions.
\item 
When completing stories from the dataset, the completions usually turn out to be very different than the original story.
\item 
Typical $k$-grams in generated completions rarely appear in the dataset, for values of $k$ as small as $4$ or $5$.
\item 
The closest point in the dataset to each generated completion is typically still quite far from it.
\end{itemize}

All the above, taken together with the ability of models trained on \TNI~to successfully follow sets instructions which we can easily be verified to be disjoint from the dataset (for example, combinations of words can be checked), provides strong evidence that our models produce genuinely novel and diverse stories, rather than simple variations of existing stories. 

We remark that nevertheless, we are not able to completely rule out the possibility that the models perform complex template matching, as it is hard to define and measure what constitutes a novel plot or a novel story. We acknowledge that this is a limitation of our evaluation. Another possibility is that the stories in the dataset essentially span the entirety of support of the distribution in the (weak) metric of complex template matching.

\section{Interpretability}
Understanding the inner workings of deep neural networks and language models in particular is a major challenge in this field of study. For example, it is often difficult to assign a specific function to a given component of a neural network. This may be because, contrary to our intuition based on human-designed programs, the network components may not have distinct roles, but rather interact in a complex and messy way. In this section, we present some preliminary evidence that training smaller models on \TN~leads to higher interpretability, suggesting that when networks are constrained in size, we may be able to gain some insights into their internal mechanisms. We focus on two aspects of the model: the attention heads and the neurons in the MLP. 

As this is not the main focus on our paper, this section is by no means exhaustive and much more work is required in order to reach more conclusive findings. Rather, we only give some preliminary evidence which may hopefully motivate future work.

\paragraph{Attention heads.} In the study of attention heads, we take advantage of the fact that we were able to train a very shallow model (having only one transformer block) which still manages to generate meaningful text. Since the model has only one layer, the attention heads are directly responsible for generating the output tokens, and thus they may have more interpretable functions than in deeper models. We use the method of Voita et al~\cite{voita2019analyzing} to analyze the attention patterns of the heads and classify them into different types, such as positional, syntactic, or semantic. We also use the method of Clark et al~\cite{clark2019does} to visualize the attention maps of the heads and inspect their behavior on specific examples. 

Our findings suggest that the attention heads exhibit diverse and meaningful functions, such as attending to the previous word, the subject of the sentence, the end of the sentence, or the main topic of the story. We also observe that some attention heads specialize in generating certain types of words, such as nouns, verbs, or punctuation. These results suggest that the attention heads learn to perform different linguistic tasks and capture different aspects of the stories.

\paragraph{Neurons in the MLP.} We also give some initial evidence that in smaller models, some neurons in the MLP have roles that are interpretable by humans. We use the method similar to~\cite{li2015visualizing} to identify the most influential tokens in the MLP for each neuron. We find that some neurons are activated on words that have a specific role in the sentence (such as the subject or the action), or in the story (such as the introduction of the protagonist). These findings suggest that the neurons in the MLP learn to encode different semantic and stylistic information and influence the generation process.

\subsection{Interpreting the role of different attention heads}
To understand the model's attention pattern after training, we use a 1-layer model with hidden dimension 1024 and 16 attention heads that was trained on \TN. We visualize the attention patterns that it produces when processing the following paragraph (the bold form is the prompt, the highlighted text is generated by the model):

\begin{quote}
\scriptsize
\textbf{One day, Lucy asks Tom: "I am looking for a banana but I can't find it". Tom says: "Don't} \hl{worry, I will help you". Lucy and Tom go to the park. They look for the banana together. After a while, they found the banana. Lucy is happy. She says: "Thank you, Tom. You are a good friend." Tom: "You are welcome, Lucy. I am happy to help you. Let's eat the banana together!"}
\end{quote}

There seems to be a clear separation between heads with attention pattern based mainly on the distance between tokens, and heads whose attention pattern has a stronger dependence on the semantic meaning:

\paragraph{Distance based attention.} Out of the 16 attention heads, we observe multiple positional-based attention heads, such that each token attends to tokens with a prescribed relative distance. Different heads are associated with different distances.

\begin{figure}[h!]
\centering
\includegraphics[width=0.22\textwidth]{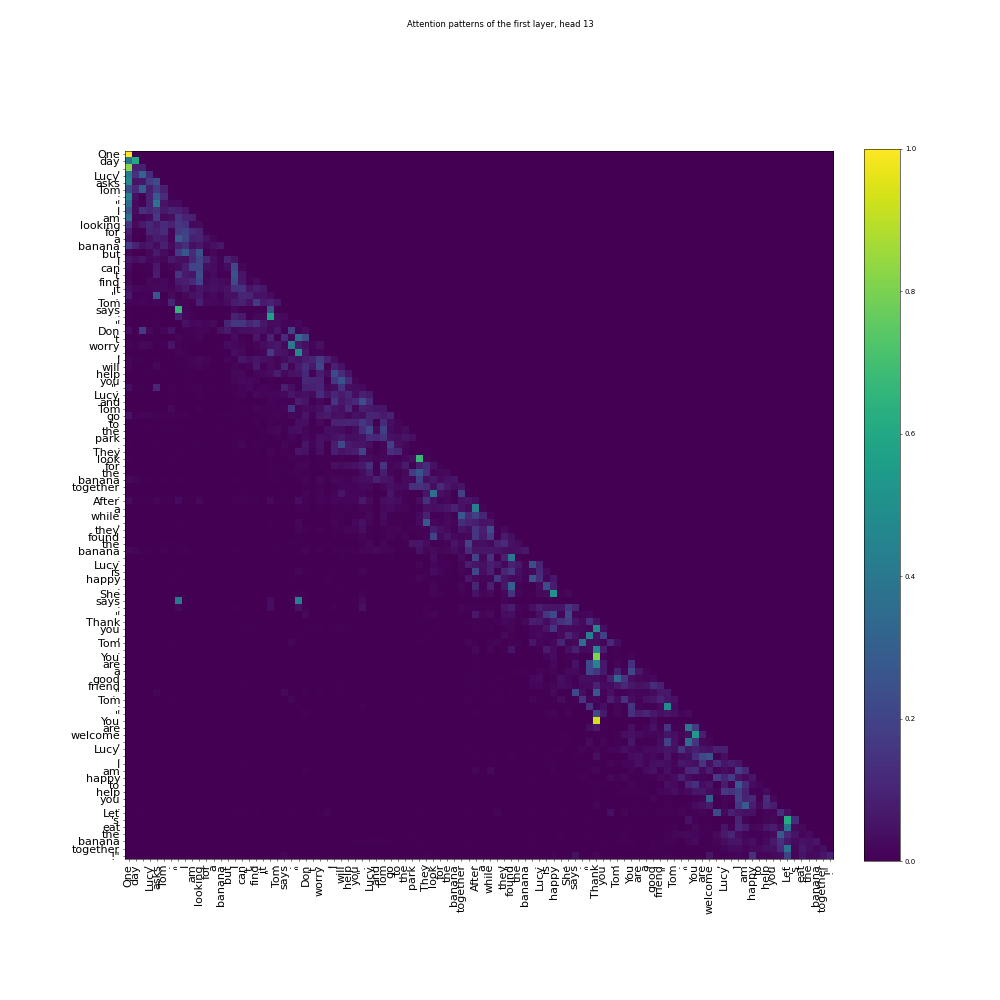}
\includegraphics[width=0.22\textwidth]{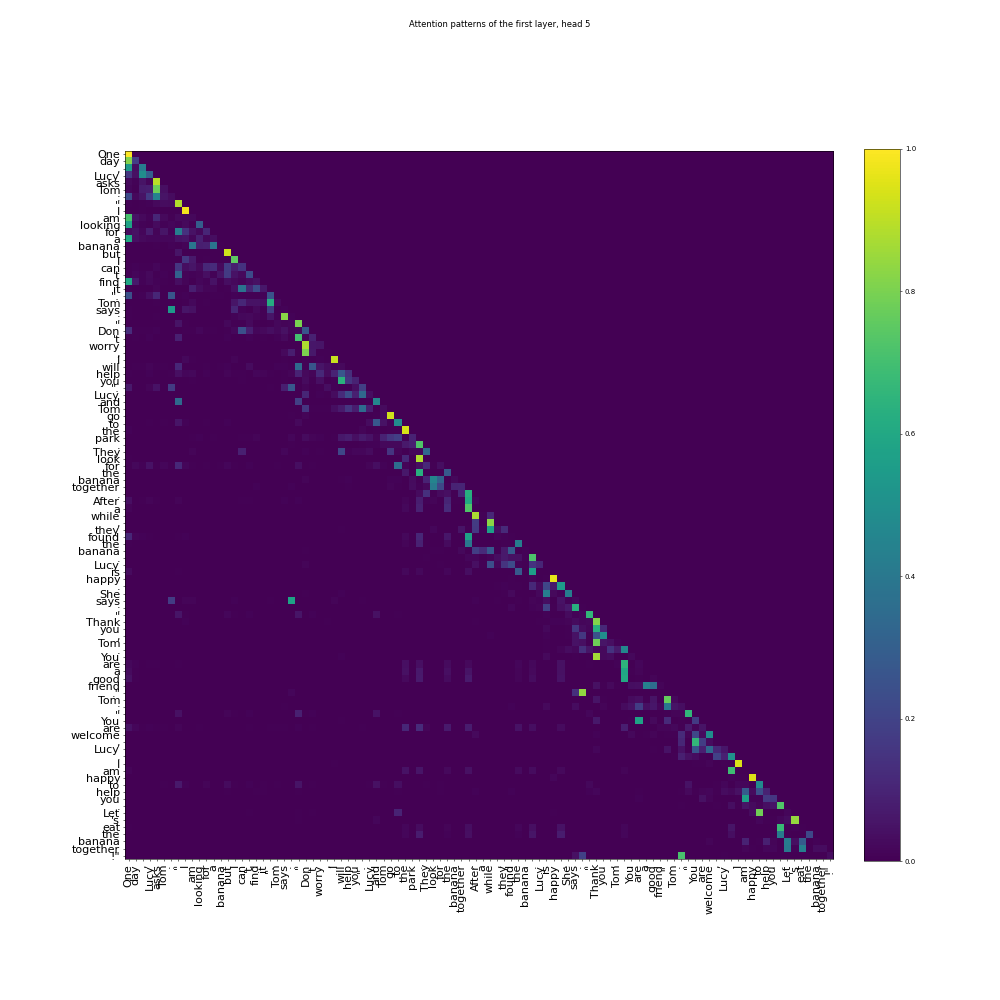}
\includegraphics[width=0.22\textwidth]{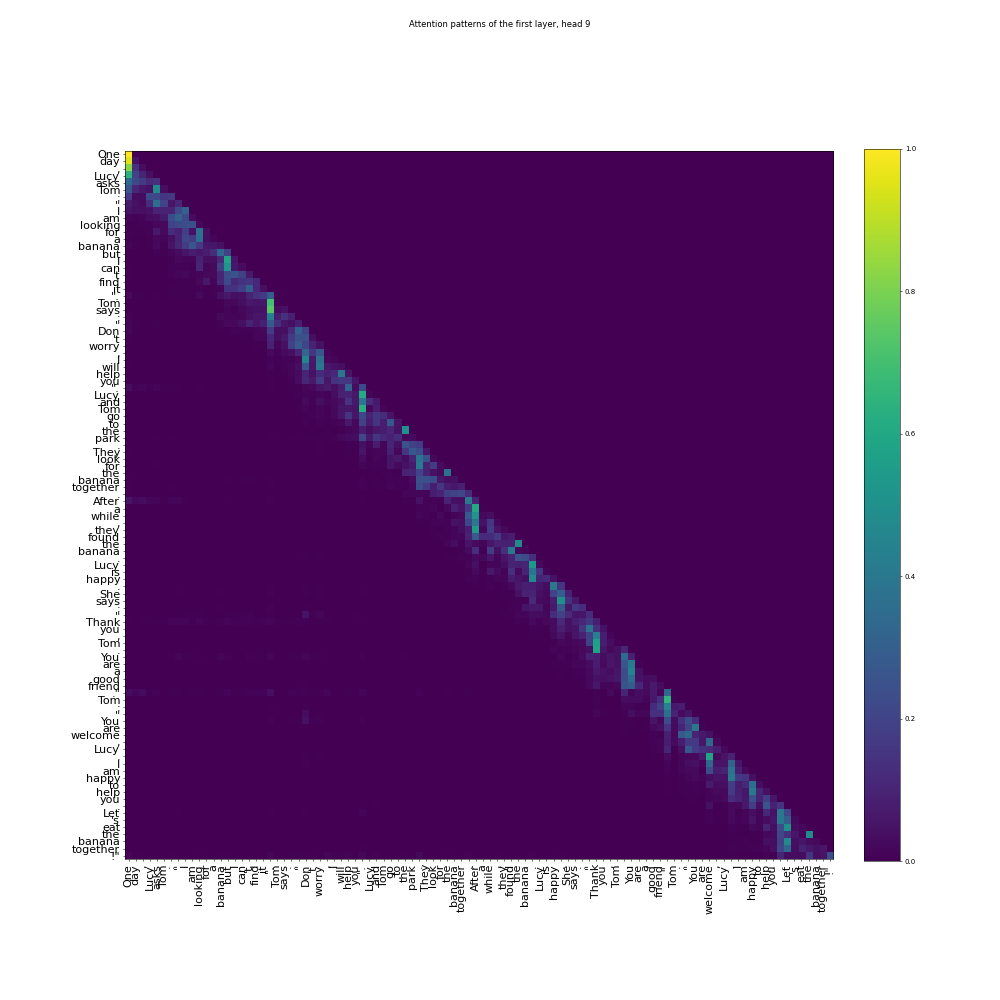}
\includegraphics[width=0.22\textwidth]{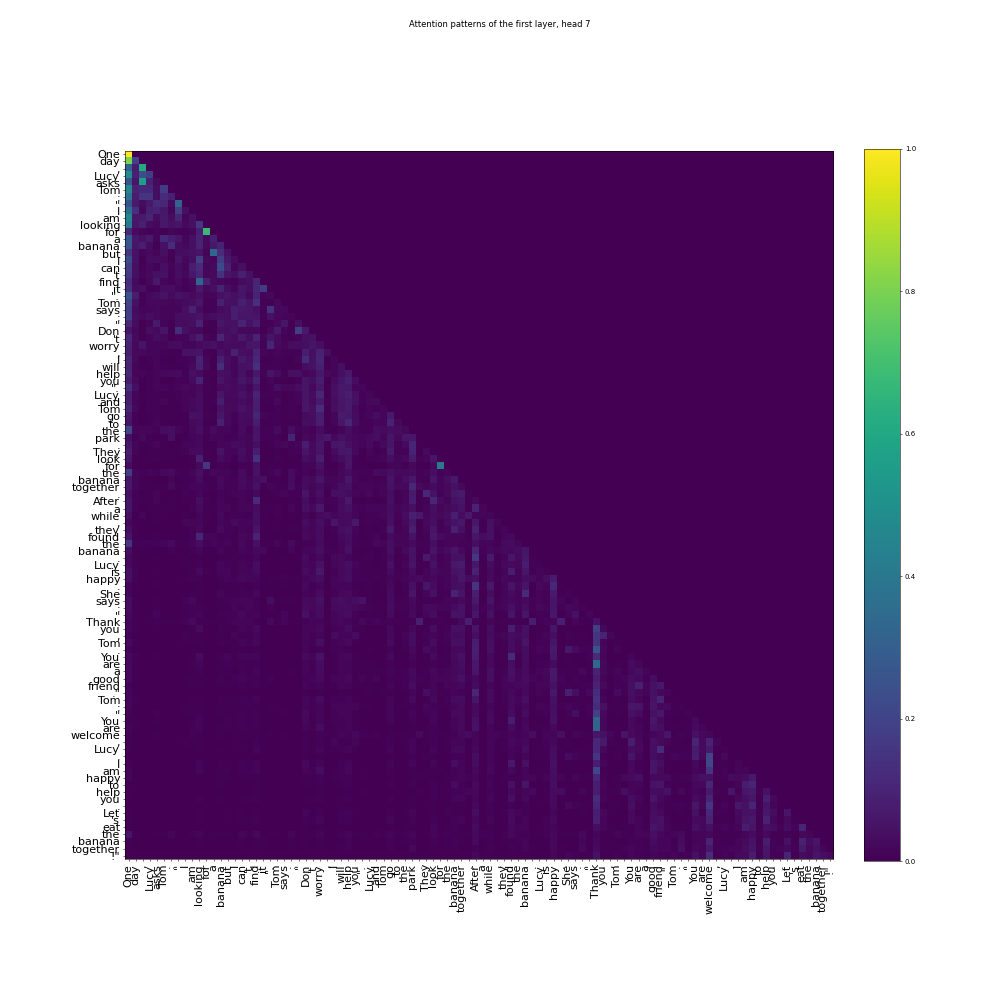}
\includegraphics[width=0.22\textwidth]{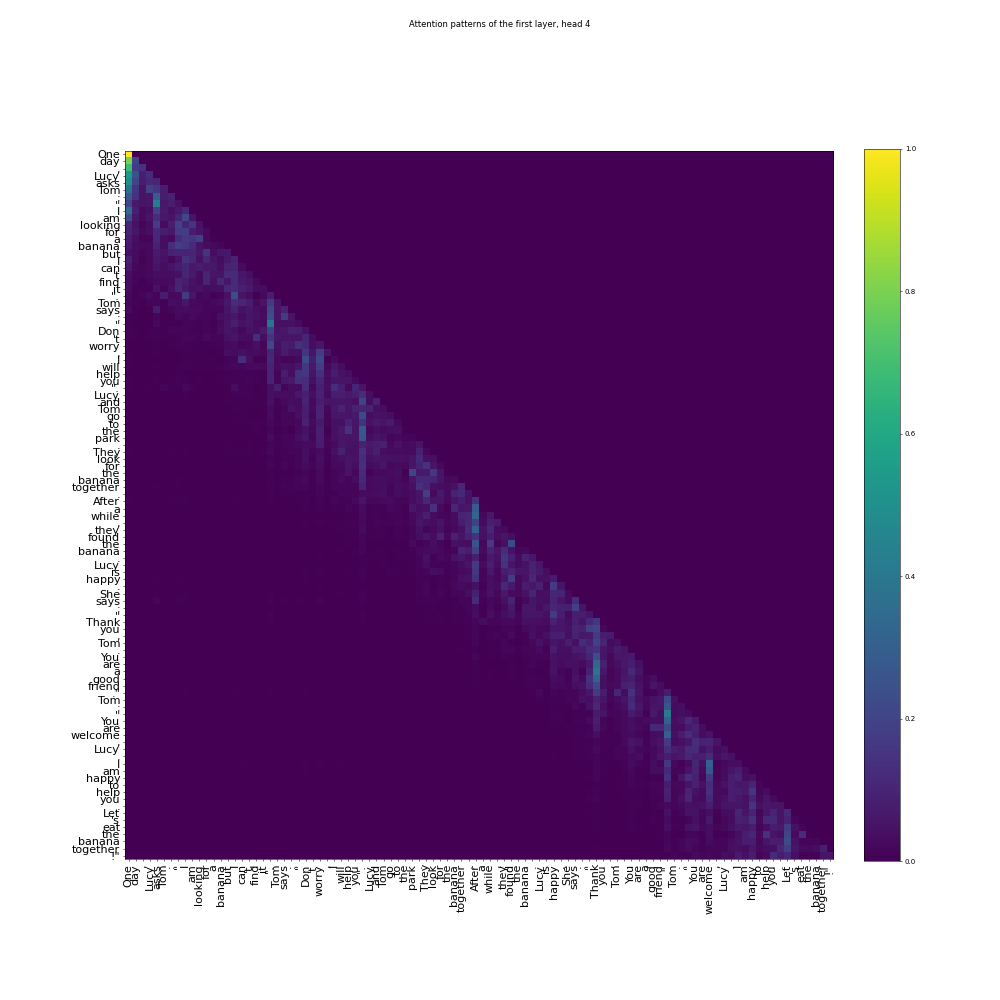}
\includegraphics[width=0.22\textwidth]{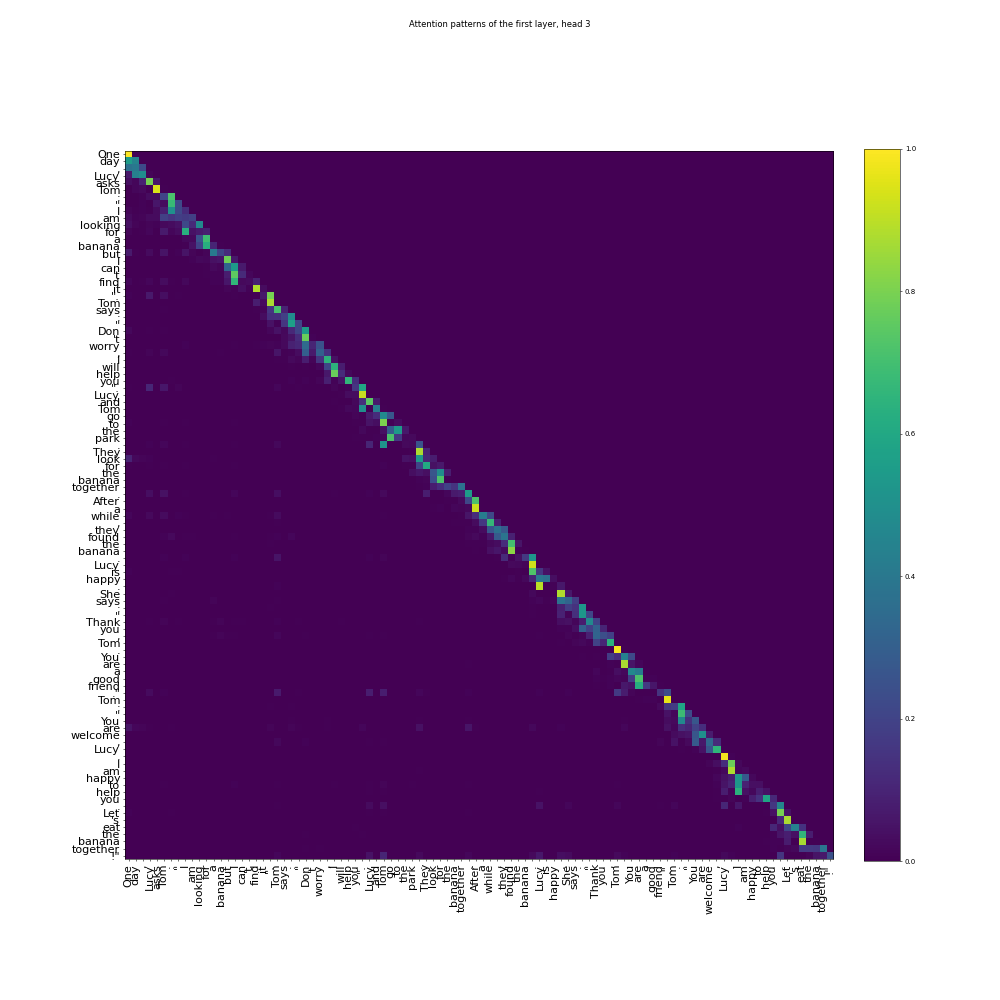}
\includegraphics[width=0.22\textwidth]{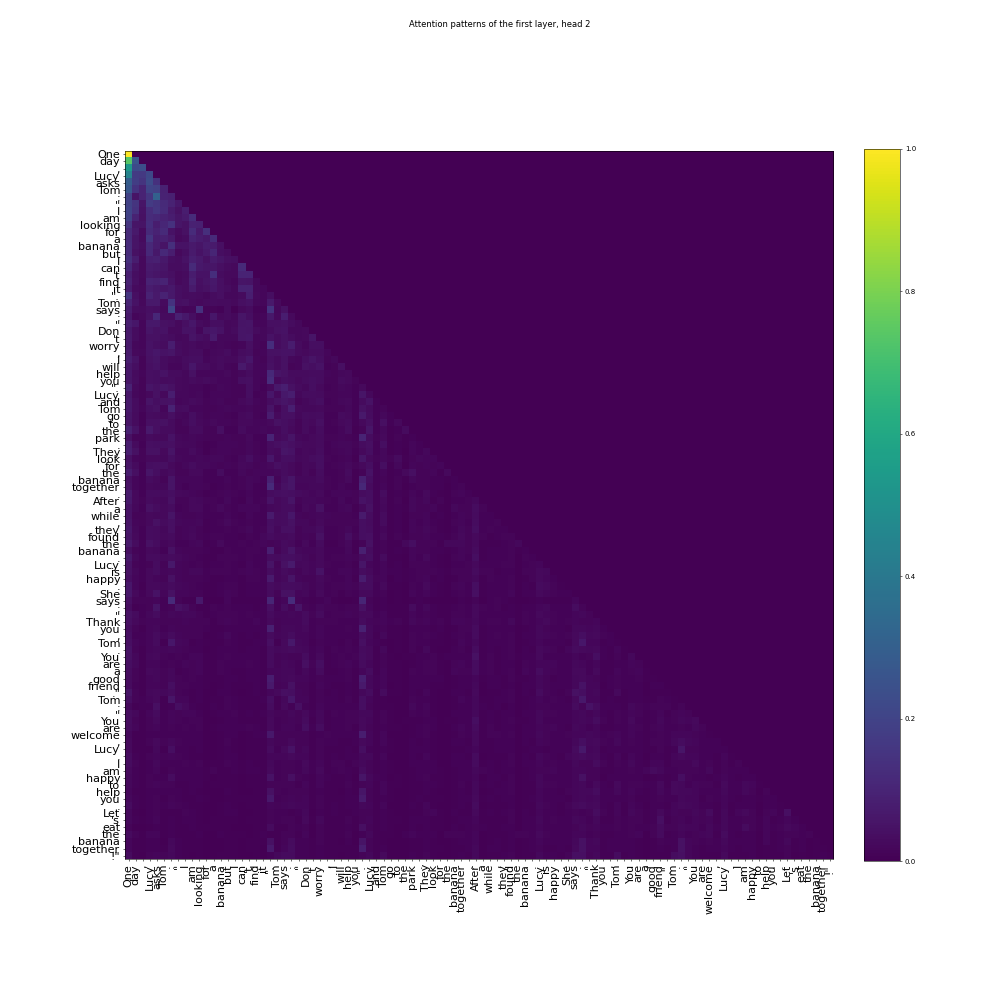}
\caption{Multi-scale distance-based attention.}
\label{fig:singleline1}
\end{figure}

\paragraph{Semantic based attention.} We also observe that there is (1). one head that the word ``the'' and ``a'' all attend to the word ``banana'', interestingly, the ``the'' at ``the park'' also attends to ``banana'', but the model still manage to generate ``park'', which is the consistent completion. (2). Another attention head gives a pattern where the tokens ``the'' and ``a'' all attend to ``park''. (3). There is third head that most of the words attend to the name of ``Tom'' and ``Lucy''. 

We remark that it makes sense that the generation of words like ``the'', ``a'', ``and'' or ``,'' would be induced by distance-based, \emph{local} attention heads, since those are tokens with a grammatical role which depends on the short-range interactions within a single sentence. On the other hand, the main entities in the story such as ``banana'', ``park'', ``Lucy'' and ``Tom'' cannot usually be predicted (as a next token) only based on the neighboring tokens, which is why the model needs to use semantic attention heads for their generation.

\begin{figure}[h!]
\centering
\includegraphics[width=0.30\textwidth]{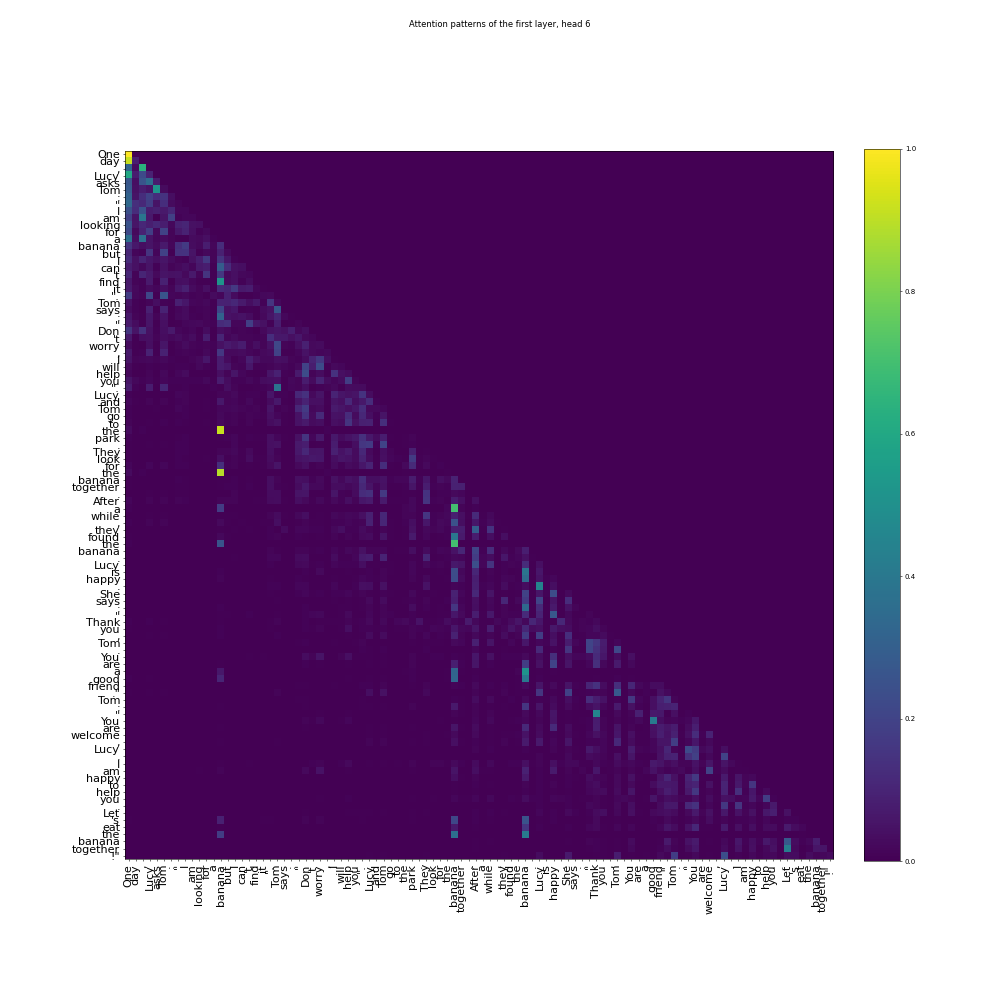}
\includegraphics[width=0.30\textwidth]{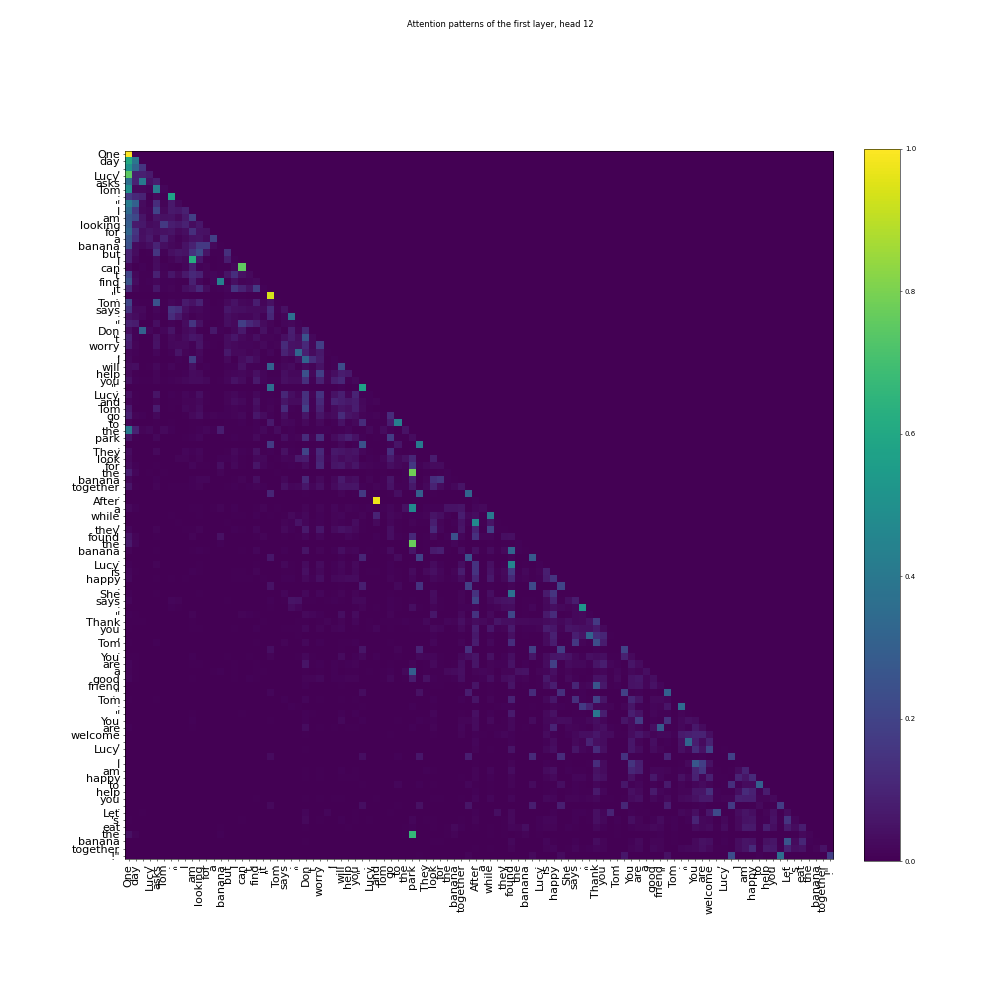}
\includegraphics[width=0.30\textwidth]{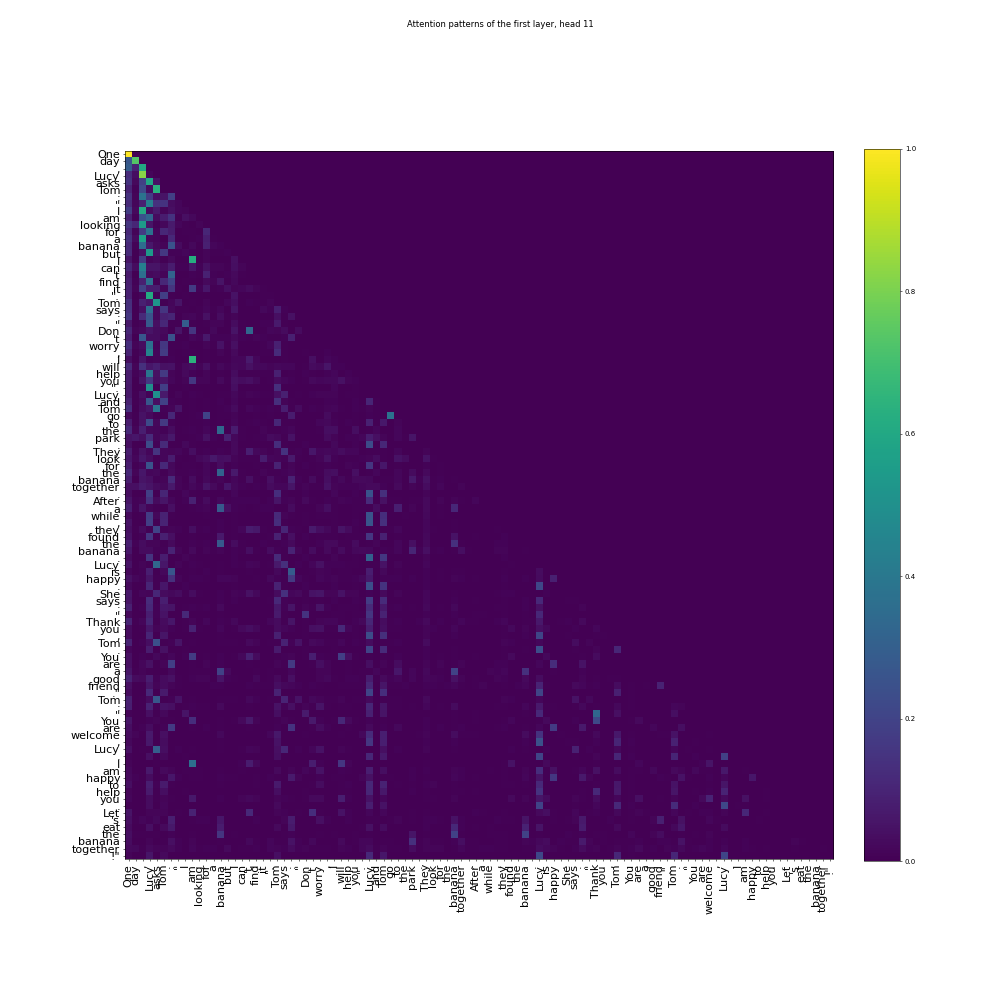}
\caption{Semantic attentions according to (1), (2), (3).} 
\label{fig:singleline2}
\end{figure}

\subsection{Interpreting the roles of different Neurons}
In order to examine whether neurons have meaningful roles, we follow \cite{li2015visualizing}, and visualize the most significant tokens for each neuron. More precisely, we take a collection of 20 stories (about 8,000 tokens) from our dataset. We take a model that was trained on \TN, we pick a transformer layer, and from the MLP associated with it we pick one coordinate in its intermediate layer. We refer to such a choice as a \emph{neuron}. We process the collection of stories with the model to obtain their internal representations, which gives us an activation value for each combination of token and neuron. Then, for each neuron we look at the tokens with highest activations from the entire collection. We highlight those tokens in red (and present them along with the sentence they are contained in). We repeated this for two models: a small model of hidden dimension 64 and 1M parameters, trained on \TN~(Figure \ref{fig:neurons64}), and on GPT2-XL (Figure \ref{fig:neuronsxl}).

In the 1M-parameter model trained on \TN, Figure \ref{fig:neurons64} first presents the activated tokens for the first two neurons in the before-last layer\footnote{The rationale behind choosing the penultimate layer is that tokens have already been processed by most layers at this point. We take the before-last rather than the last layer since the hidden representation in the last layer only has only the role of predicting the next token, so information may be lost at that point.}. Note that, since the architecture is invariant to permutations between neurons, taking the two first neurons is the same as taking an arbitrary choice of two neurons, the point being that these neurons \textbf{are neither unique nor have been cherry-picked}. We see (top row of the figure) that each of those neurons is activated on tokens with a common role (one is activated on pronouns which are also the subject in the sentence, and the other is activated on the action in the sentence). In addition, we present the activated tokens for the first neuron in another layer (layer 6), where the neuron is activates only on adjectives. Finally, we picked the neuron which has the largest activation values over all combinations of token and neuron. This neuron (depicted in the bottom right) seems to have the role of identifying the first time that the protagonist of the story is presented.

For comparison, Figure \ref{fig:neuronsxl} presents the activated tokens for first two neurons of layer 12 for GPT-XL, a much larger neural network. In this case, none of the two neurons seem to have an apparent role.

\begin{figure}[h!]
\centering
\begin{minipage}{0.45\textwidth}
\centering
\fbox{ 
\begin{minipage}{0.9\textwidth}
\scriptsize
\begin{center}
\textbf{Layer \#7, Neuron \#1} 
\end{center}
\vspace{0.1em}
Mom and Dad smiled and said, "\red{We} thought you would like this better, Lily!" \vspace{0.5\baselineskip}

Tim said, "\red{I} know it's yummy, but I don't want my tummy to hurt. \vspace{0.5\baselineskip}

Her mom said, "\red{I} don't know, Lucy. \vspace{0.5\baselineskip}

Dad said, "\red{I} turned off the water to fix a pipe. \vspace{0.5\baselineskip}

Sam thought for a moment and said, "\red{I} think I left it in the kitchen." \vspace{0.5\baselineskip}

Sam said, "\red{I}'m sorry I lost it.
\end{minipage}
}
\end{minipage}
\hfill 
\begin{minipage}{0.45\textwidth}
\centering
\fbox{
\begin{minipage}{0.9\textwidth}
\scriptsize
\begin{center}
\textbf{Layer \#7, Neuron \#2} 
\end{center}
\vspace{0.1em}
The bird flew up to the tree and tried to \red{push} the ball out.\vspace{0.5\baselineskip}

She kicked it and \red{ran} after it, laughing.\vspace{0.5\baselineskip}

She \red{pushed} and \red{pulled}, but the box would not open.\vspace{0.5\baselineskip}

They both pushed and \red{pulled}, but the tough box still did not open.\vspace{0.5\baselineskip}

Then, she saw her friend Tom \red{come} to the park.\vspace{0.5\baselineskip}

She found her toy box and \red{pushed} it to the shelf.
\end{minipage}
}
\end{minipage}

\vspace{1em} 

\begin{minipage}{0.45\textwidth}
\centering
\fbox{
\begin{minipage}{0.9\textwidth}
\scriptsize
\begin{center}
\textbf{Layer \#6, Neuron \#1} 
\end{center}
\vspace{0.1em}
They went home and shared the \red{delicious} apple. \vspace{0.5\baselineskip}

She did not like the \red{mean} king. \vspace{0.5\baselineskip}

The duck did not like the sm \red{elly} pond. \vspace{0.5\baselineskip}

The \red{new} pond was not smelly. \vspace{0.5\baselineskip}

Lucy loved to play outside under the \red{big} sky. \vspace{0.5\baselineskip}

He suggested, "Let's play a game to forget the \red{scary} wind."
\end{minipage}
}
\end{minipage}
\hfill
\begin{minipage}{0.45\textwidth}
\centering
\fbox{
\begin{minipage}{0.9\textwidth}
\scriptsize
\begin{center}
\textbf{Layer \#7, Neuron \#54} 
\end{center}
\vspace{0.1em}
One day, a girl named \red{Amy} wanted to have a fun day with her friends.\vspace{0.5\baselineskip}

Once upon a time, there was a modest girl named \red{Sue}.\vspace{0.5\baselineskip}

On the mountain, there was a small boy named \red{Tim}.\vspace{0.5\baselineskip}

One day, a girl named \red{Sue} found a big, tough box. \vspace{0.5\baselineskip}

Once upon a time, in an ancient land, there lived a little frog named \red{Freddy}.
\end{minipage}
}
\end{minipage}
\caption{Tokens which induce high activations to different neurons, for a small model trained on \TN.}
\label{fig:neurons64}
\end{figure}

\begin{figure}[h!]
\centering
\label{fig:nn_outputs512}

\begin{minipage}{0.45\textwidth}

\centering
\fbox{ 
\begin{minipage}{0.9\textwidth}
\scriptsize
\begin{center}
\textbf{Layer \#13, Neuron \#1} 
\end{center}
\vspace{0.1em}
It's not safe \red{to} play in the fog. \vspace{0.5\baselineskip}

She liked to \red{keep} her toys and books in the right place.\vspace{0.5\baselineskip}

Once upon a time, there was a gr \red{umpy} nurse.\vspace{0.5\baselineskip}

She was \red{quiet} and kind.\vspace{0.5\baselineskip}

The gr \red{umpy} nurse liked Lily very much.\vspace{0.5\baselineskip}

She called her friends on the \red{phone} and said, "Hi! \vspace{0.5\baselineskip}

Sally was \red{tired} from playing, so she went inside.\vspace{0.5\baselineskip}

Lucy was very \red{upset}.
\end{minipage}
}
\end{minipage}
\hfill 
\begin{minipage}{0.45\textwidth}
\centering
\fbox{
\begin{minipage}{0.9\textwidth}
\scriptsize
\begin{center}
\textbf{Layer \#12, Neuron \#2} 
\end{center}
\vspace{0.1em}
Mia looked at Worry and said, "My dad is in the \red{navy}.\vspace{0.5\baselineskip}

So, the \red{wise} fish told the little fish to mark a spot in the sea where they could meet every day.\vspace{0.5\baselineskip}

The \red{wise} fish told the little fish, "Always remember the spot you mark and never forget to learn new things."\vspace{0.5\baselineskip}

Suddenly, the \red{dinosaurs} came to life!
But the \red{dinosaurs} were nice and just wanted to play.\vspace{0.5\baselineskip}

He was stuck in the mirror \red{world} with Jim and could not go back to his mom and dad.
\vspace{0.1em}

\end{minipage}
}
\end{minipage}
\caption{Tokens which induce high activations to different neurons in \textbf{GPT-XL}} \label{fig:neuronsxl}
\end{figure}

\section{Exploring architectures and hyperparameters for NLP with \TN}

One of the main challenges in developing large language models (LLMs) comes from the high computational cost involved in training. Finding the best architectures, training algorithms and hyperparameters for LLMs requires a lot of resources and experimentation. Therefore, it would be useful to have a smaller and simpler dataset that can still capture some of the basic capabilities of LLMs, and allow us to study how different design choices affect their performance. \TN~is such a dataset, as it enables us to train and evaluate LMs that are orders of magnitude smaller than the state-of-the-art models, yet still have the basic capability of producing coherent text.

In this work, we take the first steps towards using \TN~as a testbed for exploring architectures and hyperparameters for NLP. We show that our small models exhibit some similar patterns to the ones observed in LLMs in certain aspects. In particular, we investigate two questions: how to balance model size and learning budget for a fixed amount of training flops, and how to choose the number of attention heads for a given model width and depth.

\paragraph{Model size versus the training FLOPs.} For a fixed amount of training flops, there is a trade-off between the size of the model and the number of training steps (the total number of flops is the product of both). Previous works \cite{kaplan2020scaling, hoffmann2022training} have shown that there is a polynomial scaling law between model size and learning budget for LLMs, i.e., the optimal model size for a given amount of flops is proportional to the flops raised to some power $\alpha > 1$. However, these works used different ranges of model sizes (from a few million to tens of billions of parameters) and found different values of $\alpha$ (around 0.7 and 0.5, respectively). A natural question is whether this scaling law is universal or depends on the dataset. Our dataset allows us to conduct a similar experiment but with much smaller models and flops. Surprisingly, we find evidence for a polynomial scaling law as well, which suggests that there might be a universal phenomenon here.

\begin{figure}[h!]\label{fig:flop}
\includegraphics[width=1\textwidth]{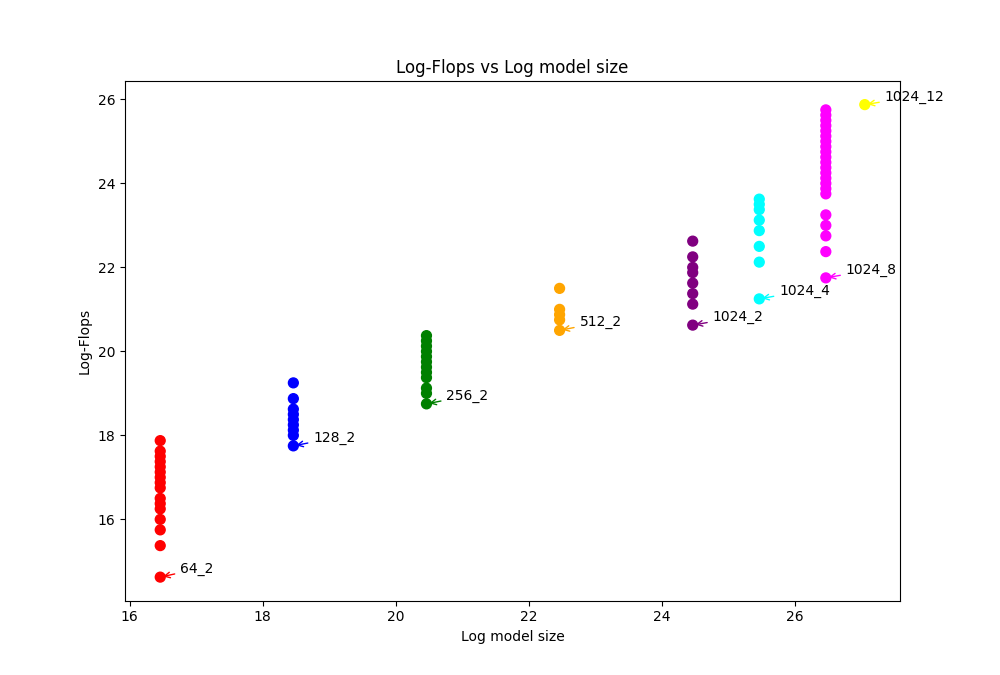}
\caption{The scaling law of the best model versus the total number of training flops.}
\end{figure}

We train models of various sizes and architectures on \TN. For each amount of flops, we select the model and the number of training steps that achieve the lowest validation loss among the possible combinations. We vary the number of layers from $2, 4, 8, 12$ and the hidden dimension from $64, 128, 256, 512, 768, 1024, 2048$. The result is shown in Figure~\ref{fig:flop}. Although the number of points may be a bit small for the data to be very conclusive, the plot points to a polynomial dependence.

\paragraph{Choosing the number of heads.} Another design choice for transformers is the number of attention heads for each layer. It is not obvious how the number of heads affects the performance of the model, given a fixed model width and depth. Our results, shown in Figure \ref{tb:1}, suggest that in the regime where the number of heads is small, increasing it improves the performance of the model across all metrics.

\begin{figure}[h!]
\centering
\begin{tabular}{ccccccc}
\hline
Hidden size & Layer & Head & Eval loss & Grammar & Creativity & Consistency \\
\hline
768 & 2 & 2 & 1.38 & 7.77 & 6.5 & 7.78 \\
768 & 2 & 4 & 1.34 & 8.05 & 6.57 & 8.16 \\
768 & 2 & 8 & 1.33 & 8.25 & 6.53 & 8.16 \\
768 & 1 & 2 & 1.58 & 7.13 & 5.83 & 6.38 \\
768 & 1 & 4 & 1.56 & 7.43 & 5.90 & 6.75 \\
768 & 1 & 8 & 1.54 & 7.45 & 6.28 & 7.02 \\
\hline
\end{tabular}
\caption{Model performance with different number of attention heads} \label{tb:1}
\end{figure}

\section{Related Works}
Generative language models (LMs) have achieved impressive results in various natural language processing tasks, such as text summarization, dialogue generation, and story completion. However, most of these models are very large, with hundreds of millions or even billions of parameters, which poses significant challenges for training, inference, and deployment. For example, GPT-3~\cite{brown2020language}, one of largest LM to date, has 175 billion parameters and requires hundreds of petaflops of compute to train. Smaller models, such as GPT-2 small with 125 million parameters, can hardly generate coherent and consistent sentences beyond a few words, even after extensive pre-training on large corpora~\cite{radford2019language}.

Several methods have been proposed to compress or distill large LMs into smaller ones, such as knowledge distillation~\cite{hinton2015distilling,allen2020towards}, pruning~\cite{frankle2018lottery}, and quantization~\cite{hubara2017quantized}. However, these methods are much more effective for BERT-like models~\cite{sanh2019distilbert,sun2020mobilebert}, which are designed for masked language modeling and downstream classification tasks, than for GPT-like models, which are designed for autoregressive language generation~\cite{santacroce2023matters}. 

Another challenge for generative LMs is the evaluation of their outputs. Unlike BERT-like models, which can be fine-tuned and evaluated on downstream tasks with labeled data, GPT-like models are more difficult to measure in terms of how well they can "speak and understand natural language". Most existing benchmarks for generative LMs, such as LAMBADA~\cite{paperno2016lambada}, CLOZE ~\cite{taylor1953cloze}, TriviaQA~\cite{joshi2017triviaqa}, and Winograd Schema Challenge~\cite{levesque2012winograd}, require the models to produce a single word or a short phrase as the answer, which does not capture the richness and diversity of generating natural language. Moreover, these benchmarks are often limited by the size and quality of the datasets, the ambiguity and subjectivity of the answers, and the lack of human evaluation. Larger and more diversed datasets such as the BigBench~\cite{srivastava2022beyond} are simply way too complicated for SLMs. Some other benchmarks, such as WikiSQL~\cite{zhong2017seq2sql}, have a more structured output format, which makes them easier to evaluate, but also less representative of natural language generation. 

Our work is also beneficial to the theoretical analysis of transformer models and their learning process. Most of the existing theory works focus on models with one transformer block, which are easier to analyze than models with multiple blocks. For example, Voita et al~\cite{voita2019analyzing} showed that one transformer block can learn to perform different linguistic tasks depending on the position of the self-attention layer. Li et al~\cite{li2023transformers} shows a transformer block can encode topical models. Jelassi et al~\cite{jelassi2022vision} shows one transformer block can encode patch associations. Our work provides empirical evidence that one transformer block can also generate diverse and consistent stories, which suggests that the transformer architecture has a strong expressive power even with a small number of parameters and layers.

\section{Conclusion}
In this work, we have presented \TN, a synthetic dataset of short stories that only contain words that a typical 3 to 4-year-olds usually understand, generated by GPT-3.5 and GPT-4. We have shown that \TN~can be used to train and evaluate small language models (SLMs) that are much smaller than the state-of-the-art models, yet still produce fluent and consistent stories with several paragraphs that are diverse and have almost perfect grammar, and demonstrate reasoning capabilities.

While large models trained on the huge and diverse language corpuses on the internet exhibit very impressive capabilities, those datasets appear to be too large for SLMs to capture the complex aspects of language. In this work we have argued that \TN~enables us to observe and study the emergence of capabilities such as generation of coherent text, reasoning and instruction following in LMs on a much smaller scale, in terms of the size of both model and dataset. By training SLMs on our dataset, we have also observed many behaviors similar to LLMs such as scaling laws, trade-offs between width and depth, etc. Moreover, we have shown that the trained SLMs have much higher interpretability than larger ones, and that we can visualize and analyze their attention and activation patterns to understand how they generate and comprehend stories.

We provided evidence to the fact that the models trained on \TN~are able to produce genuinely new stories, rather than just copying chunks of text the dataset. It remains a challenge, however, to assess the true extent of the "creativity" of our models, and to which the models reflect a certain "understanding" (on a very low level of course) of the stories that they produce as opposed to just template matching to create a plausible continuation. We hope that this dataset can be used in future works to obtain insights about the degree of creativity of language models.

We have also introduced a new paradigm for the evaluation of language models, which uses GPT-4 to grade the content generated by these models as if those were stories written by students and graded by a (human) teacher. This new paradigm overcomes the flaws of standard benchmarks, which often require the model's output to be very structured, and moreover provides a multidimensional score for the model, providing scores for different capabilities. We believe that this paradigm can be useful much beyond \TN.

Finally, we have presented initial findings which point to the roles of width vs. depth in the intellectual capabilities of generative networks, which suggest that width is more important for capturing factual knowledge whereas depth is more important for contextual tracking. Moreover our findings suggest that in terms of emergence, grammatic and syntactic abilities appear earlier than the ability to produce consistent text, which in turn appears ahead of ability to generate content that would be considered as creative. These preliminary findings are only suggestive (and have not been the main focus of this work) but they show how our dataset and evaluation paradigm can enable more fine-grained analysis of the emergence and evaluation of various language capabilities in generative models.

We hope that \TN~can facilitate the development, analysis and research of LMs, especially for low-resource or specialized domains, and shed light on the emergence of language capabilities in LMs. A general question that arises from this work is whether synthesizing a refined dataset can be beneficial in training networks for practical uses. For example, perhaps it is possible to train a customer service chatbot by synthesizing a large dataset of hypothetical calls.

\bibliographystyle{plain}
\bibliography{main}
\end{document}